\documentclass{article} 
\usepackage{iclr2023_conference,times}

\usepackage{amsmath,amsfonts,bm}

\def\eqref#1{equation~\ref{#1}}

\def\1{\bm{1}}

\DeclareMathAlphabet{\mathsfit}{\encodingdefault}{\sfdefault}{m}{sl}
\SetMathAlphabet{\mathsfit}{bold}{\encodingdefault}{\sfdefault}{bx}{n}

\usepackage{hyperref}
\usepackage{url}
\usepackage{tabularx}
\usepackage{graphicx}
\usepackage{caption}
\usepackage{subcaption}
\usepackage{amsmath}
\usepackage{float}
\usepackage[linesnumbered,ruled,vlined]{algorithm2e}
\usepackage{booktabs}
\usepackage{amsfonts}
\usepackage{amsmath}
\usepackage{amssymb}
\usepackage{authblk}

\newcommand{\dataset}{\mathcal{D}}
\newcommand{\traj}{\tau}
\newcommand{\numtraj}{n_{\traj}}
\newcommand{\obs}{o}
\newcommand{\ac}{a}
\newcommand{\rew}{r}
\newcommand{\mathbbm}[1]{\text{\usefont{U}{bbm}{m}{n}#1}} 

\title{Explaining RL Decisions with Trajectories}

\author[1]{\textbf{Shripad Deshmukh}$^{*,}$}
\author[2]{\textbf{Arpan Dasgupta}$^{\ddagger,}$}
\author[1]{\textbf{Balaji Krishnamurthy}}
\author[3]{\textbf{Nan Jiang}}
\author[1]{\textbf{Chirag Agarwal}}
\author[4]{\textbf{Georgios Theocharous}}
\author[1]{\textbf{Jayakumar Subramanian}}
\affil[1]{Media and Data Science Research, Adobe}
\affil[2]{International Institute of Information Technology Hyderabad}
\affil[3]{University of Illinois Urbana-Champaign}
\affil[4]{Adobe Research}

\iclrfinalcopy 
\begin{document}

\maketitle

\let\thefootnote\relax\footnotetext{$^{*}$Email for correspondence: \texttt{shdeshmu@adobe.com}}\let\thefootnote\relax\footnotetext{$^{\ddagger}$ Work done during summer research internship at Media and Data Science Research, Adobe.}

\begin{abstract}
    Explanation is a key component for the adoption of reinforcement learning (RL) in many real-world decision-making problems. In the literature,  the explanation is often provided by saliency attribution to the features of the RL agent's state. In this work, we propose a complementary approach to these explanations, particularly for offline RL, where we attribute the policy decisions of a trained RL agent to the trajectories encountered by it during training. To do so, we encode trajectories in offline training data individually as well as collectively (encoding a set of trajectories). We then attribute policy decisions to a set of trajectories in this encoded space by estimating the sensitivity of the decision with respect to that set.  Further, we demonstrate the effectiveness of the proposed approach in terms of quality of attributions as well as practical scalability in diverse environments that involve both discrete and continuous state and action spaces such as grid-worlds, video games (Atari) and continuous control (MuJoCo). We also conduct a human study on a simple navigation task to observe how their understanding of the task compares with data attributed for a trained RL policy. Code-base can be found 
\\ \href{https://github.com/shripaddeshmukh/xrl\_with\_trajectories}{here}. 
\end{abstract}

\section{Introduction} \label{sec:introduction}

Reinforcement learning~\citep{sutton2018reinforcement} has enjoyed great popularity and has achieved huge success, especially in the online settings, post advent of the deep reinforcement learning~\citep{mnih2013playing, schulman2017proximal, silver2017mastering, haarnoja2018soft}. Deep RL algorithms are now able to handle high-dimensional observations such as visual inputs with ease. However, using these algorithms in the real world requires - i) efficient  learning from minimal exploration to avoid catastrophic decisions due to insufficient knowledge of the environment, and ii) being explainable. The first aspect is being studied under offline RL where the agent is trained on collected experience rather than exploring directly in the environment. There is a huge body of work on offline RL~\citep{DBLP:journals/corr/abs-2005-01643, kumar2020conservative, yu2020mopo, kostrikov2021offline}. However, more work is needed to address the explainability aspect of RL decision-making.

Previously, researchers have attempted explaining decisions of RL agent by highlighting important features of the agent's state (input observation)~\citep{puri2019explain, iyer2018transparency, greydanus2018visualizing}. While these approaches are useful, we take a complementary route. Instead of identifying salient state-features, we wish to identify the past experiences (trajectories) that led the RL agent to learn certain behaviours. We call this approach as trajectory-aware RL explainability. Such explainability confers faith in the decisions suggested by the RL agent in critical scenarios (surgical~\citep{loftus2020decision}, nuclear~\citep{Boehnlein_2022}, etc.) by looking at the trajectories responsible for the decision.  While this sort of training data attribution has been shown to be highly effective in supervised learning~\citep{nguyen2021effectiveness}, to the best of our knowledge, this is the first work to study data attribution-based explainability in RL. In the present work, we restrict ourselves to offline RL setting where the agent is trained completely offline, i.e., without interacting with the environment and later deployed in the environment.

Contributions of this work are enumerated below:
\begin{enumerate}
    
    \item A novel explainability framework for reinforcement learning that aims to find experiences(trajectories) that lead an RL agent learn certain behaviour. 
    
    \item A solution for trajectory attribution in offline RL setting based on state-of-the-art sequence modeling techniques. In our solution, we present a methodology that generates a single embedding for a trajectory of states, actions, and rewards, inspired by approaches in Natural Language Processing (NLP). We also extend this method to generate a single encoding of data containing a set of trajectories.
    
    \item Analysis of trajectory explanations produced by our technique along with analysis of the trajectory embeddings generated, where we demonstrate how different embedding clusters represent different semantically meaningful behaviours. Additionally, we also conduct a study to compare human understanding of RL tasks with trajectories attributed.

\end{enumerate}

This paper is organized as follows. In Sec.~\ref{sec:relwork} we cover the works related to explainability in RL and the recent developments in offline RL. We then present our trajectory attribution algorithm in Sec.~\ref{sec:method}. The experiments and results are presented in Sec.~\ref{sec:res}. We discuss the implications of our work and its potential extensions in the concluding Sec.~\ref{sec:conc}.

\section{Background and Related Work} \label{sec:relwork}

\textbf{Explainability in RL.} Explainable AI (XAI) refers to the field of machine learning (ML) that focuses on developing tools for explaining the decisions of ML models. Explainable RL (XRL)~\citep{puiutta2020explainable, korkmaz2021investigating} is a sub-field of XAI that specializes in interpreting behaviours of RL agents. Prior works include approaches that distill the RL policy into simpler models such as decision tree~\citep{coppens2019distilling} or to human understandable high-level decision language~\citep{verma2018programmatically}. However, such policy simplification fails to approximate the behavior of complex RL models. In addition, causality-based approaches~\citep{pawlowski2020deep, madumal2020explainable} aim to explain an agent's action by identifying the cause behind it using counterfactual samples.
Further, saliency-based methods using input feature gradients~\citep{iyer2018transparency} and perturbations~\citep{puri2019explain, greydanus2018visualizing} provide state-based explanations that aid humans in understanding the agent's actions. To the best of our knowledge, for the first time, we explore the direction of explaining an agent's behaviour by attributing its actions to past encountered trajectories rather than highlighting state features. Also, memory understanding~\citep{koul2018learning, danesh2021re} is a relevant direction, where finite state representations of recurrent policy networks are analysed for interpretability. However, unlike these works, we focus on sequence embedding generation and avoid using policy networks for actual return optimization.

\textbf{Offline RL.} Offline RL~\citep{DBLP:journals/corr/abs-2005-01643} refers to the RL setting where an agent learns from collected experiences and does not have direct access to the environment during training. There are several specialized algorithms proposed for offline RL including model-free ones~\citep{kumar2020conservative, Kumar2019StabilizingOQ} and model-based ones~\citep{kidambi2020morel,yu2020mopo}. In this work, we use algorithms from both these classes to train offline RL agents. In addition, recently, the RL problem of maximizing long-term return has been cast as taking the best possible action given the sequence of past interactions in terms of states, actions, rewards~\citep{chen2021decisiontransformer,janner2021sequence,reed2022generalist, park2018sequence}. Such sequence modelling approaches to RL, especially the ones based on transformer architecture~\citep{vaswani2017attention}, have produced state-of-the-art results in various offline RL benchmarks, and offer rich latent representations to work with. However, little to no work has been done in the direction of understanding these sequence representations and their applications. In this work, we base our solution on transformer-based sequence modelling approaches to leverage their high efficiency in capturing the policy and environment dynamics of the offline RL systems. Previously, researchers in group-driven RL~\citep{zhu2018group} have employed raw state-reward vectors as trajectory representations. We believe transformer-based embeddings, given their proven capabilities, would serve as better representations than state-reward vectors.

\section{Trajectory Attribution} 
\label{sec:method}
\begin{figure}[h]
    \centering
    \includegraphics[width=0.99\linewidth]{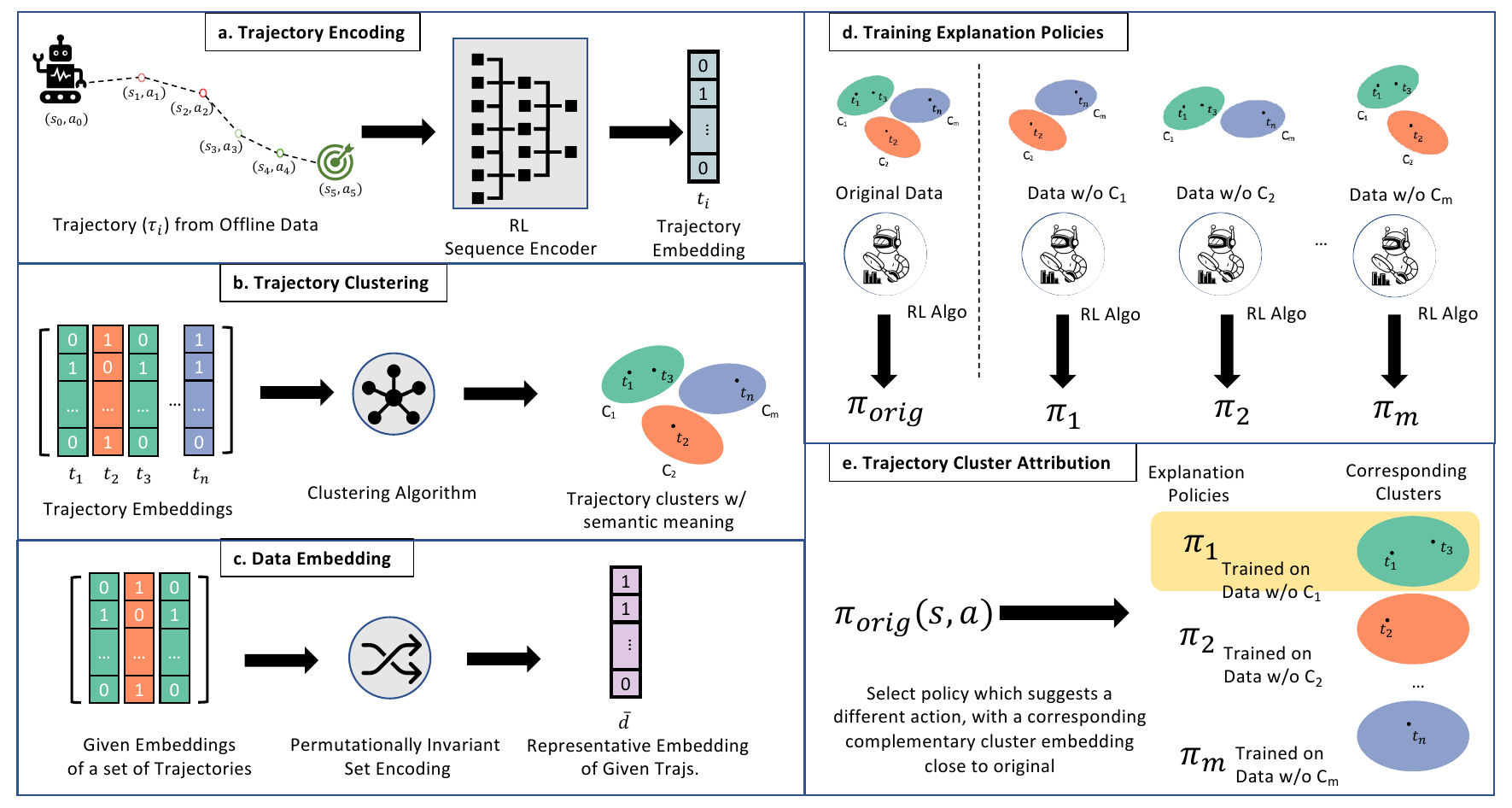}
    \caption{\textbf{Trajectory Attribution in Offline RL.} First, we encode trajectories in offline data using sequence encoders and then cluster the trajectories using these encodings. Also, we generate a single embedding for the data. Next, we train explanation policies on variants of the original dataset and compute corresponding data embeddings. Finally, we attribute decisions of RL agents trained on entire data to trajectory clusters using action and data embedding distances.}
    \label{fig:proposed_technique}
\end{figure}

\textbf{Preliminaries.} We denote the offline RL dataset using $\dataset$ that comprises a set of $\numtraj$ trajectories. Each trajectory, denoted by $\traj_j$ comprises of a sequence of observation ($\obs_k$), action ($\ac_k$) and per-step reward ($\rew_k$) tuples with $k$ ranging from 1 to the length of the trajectory $\traj_j$. We begin by training an offline RL agent on this data using any standard offline RL algorithm from the literature.

\textbf{Algorithm.} Having obtained the learned policy using an offline RL algorithm, our objective now is to attribute this policy, i.e., the action chosen by this policy, at a given state to a set of trajectories. We intend to achieve this in the following way. We want to find the smallest set of trajectories, the absence of which from the training data leads to different behavior at the state under consideration. That is, we posit that this set of trajectories contains specific behaviors and respective feedback from the environment that trains the RL agent to make decisions in a certain manner. This identified set of trajectories would then be provided as attribution for the original decision.

While this basic idea is straightforward and intuitive, it is not computationally feasible beyond a few small RL problems with discrete state and action spaces. The key requirement to scale this approach to large, continuous state and action space problems, is to group the trajectories into clusters which can then be used to analyze their role in the decision-making of the RL agent. In this work, we propose to  cluster the trajectories using trajectory embeddings produced with the help of state-of-the-art sequence modeling approaches.

Figure~\ref{fig:proposed_technique} gives an overview of our proposed approach involving five steps: (i) Trajectory encoding, (ii) Trajectory clustering, (iii) Data embedding, (iv) Training explanation policies, and (v) Cluster attribution, each of which is explained in the sequel.

\textbf{(i) Trajectory Encoding.} First, we tokenize the trajectories in the offline data according to the specifications of the sequence encoder used (e.g. decision transformer/trajectory transformer). The observation, action and reward tokens of a trajectory are then fed to the sequence encoder to produce corresponding latent representations, which we refer to as output tokens. We define the \textit{trajectory embedding} as an average of these output tokens. This technique is inspired by average-pooling techniques~\citep{choi2021evaluation, briggs_2021} in NLP used to create sentence embedding from embeddings of words present in it. (Refer to Algorithm~\ref{alg:trajectory_encoding}.)

\begin{algorithm}[H]
\footnotesize
\caption{encodeTrajectories}
\label{alg:trajectory_encoding}
\SetKwInOut{Input}{Input}
\SetKwInOut{Initialize}{Initialize}
\SetKwInOut{Output}{Output}

\tcc{Encoding given set of trajectories individually}
\vspace{1mm}
\Input{Offline Data \{$\tau_i$\}, Sequence Encoder $E$}
\Initialize{Initialize array $T$ to collect the trajectory embeddings}
\For{$\tau_j$ in $\{\tau_i\}$}{
    \tcc{Using $E$, get output tokens for all the $\obs$, $\ac$ \& $\rew$ in $\tau_j$ }
    
    $(e_{o_{1,j}}, e_{a_{1,j}}, e_{r_{1,j}}, ..., e_{o_{\mathbb{T},j}}, e_{a_{\mathbb{T},j}}, e_{r_{\mathbb{T},j}}) \gets E(o_{1,j}, a_{1,j}, r_{1,j}, ..., o_{\mathbb{T},j}, a_{\mathbb{T},j}, r_{\mathbb{T},j})$ \tcp{ where $3\mathbb{T}$ = {\#}input tokens}
    \tcc{Take mean of outputs to generate $\tau_j$'s embedding $t_j$}
    $t_j \gets (e_{o_{1,j}} + e_{a_{1,j}} + e_{r_{1,j}} + e_{o_{2,j}} + e_{a_{2,j}} + e_{r_{2,j}} + ... + e_{o_{\mathbb{T},j}} + e_{a_{\mathbb{T},j}} + e_{r_{\mathbb{T},j}}) / (3 \mathbb{T})$ 
    
    Append $t_j$ to $T$
}

\Output{Return the trajectory embeddings $T = \{t_i\}$}

\end{algorithm}

\textbf{(ii) Trajectory Clustering.} Having obtained trajectory embeddings, we cluster them using X-Means clustering algorithm~\citep{pelleg2000x} with implementation provided by~\citet{Novikov2019}. While in principle, any suitable clustering algorithm can be used here, we chose X-Means as it is a simple extension to the K-means clustering algorithm~\citep{lloyd1982least}; it determines the number of clusters $n_c$ automatically. This enables us to identify all possible  patterns in the trajectories without forcing $n_c$ as a hyperparameter (Refer to Algorithm~\ref{alg:cluster_trajs}). 

\begin{algorithm}[h!]
\footnotesize
\caption{clusterTrajectories}
\label{alg:cluster_trajs}
\SetKwInOut{Input}{Input}
\SetKwInOut{Initialize}{Initialize}
\SetKwInOut{Output}{Output}

\tcc{Clustering the trajectories using their embeddings}
\vspace{1mm}
\Input{Trajectory embeddings $T = \{t_i\}$, clusteringAlgo}

$C \gets $ clusteringAlgo(T) \tcp{Cluster using provided clustering algorithm}

\Output{Return trajectory clusters $C = \{c_i\}_{i=1}^{n_c}$}
\end{algorithm}

\textbf{(iii) Data Embedding.} We need a way to identify the least change in the original data that leads to the change in behavior of the RL agent. To achieve this, we propose a representation for data comprising the collection of trajectories. The representation has to be agnostic to the order in which trajectories are present in the collection. So, we follow the set-encoding procedure prescribed in~\citep{DBLP:journals/corr/ZaheerKRPSS17} where we first sum the embeddings of the trajectories in the collection, normalize this sum by division with a constant and further apply a non-linearity, in our case, simply, softmax over the feature dimension to generate a single \textit{data embedding} (Refer to Algorithm~\ref{alg:data_encoding}). 

We use this technique to generate data embeddings for $n_c+1$ sets of trajectories. The first set represents the entire training data whose embedding is denoted by $\Bar{d}_{\text{orig}}$. The remaining $n_c$ sets are constructed as follows. For each trajectory cluster $c_j$, we construct a set with the entire training data but the trajectories contained in $c_j$. We call this set the complementary data set corresponding to cluster $c_j$ and the corresponding data embedding as the complementary data embedding $\Bar{d}_{j}$.

\begin{algorithm}[h!]
\footnotesize
\caption{generateDataEmbedding}
\label{alg:data_encoding}
\SetKwInOut{Input}{Input}
\SetKwInOut{Initialize}{Initialize}
\SetKwInOut{Output}{Output}

\tcc{Generating data embedding for a given set of trajectories}
\vspace{1mm}
\Input{Trajectory embeddings $T = \{t_i\}$, Normalizing factor $M$, Softmax temperature $T_{\text{soft}}$}

$\Bar{s} \gets \frac{\sum_{i}t_i}{M}$ \tcp{Sum the trajectory embeddings and normalize them}

$\Bar{d} \gets \{d_j | d_j = \frac{exp({s_j /T_{\text{soft}}})}{\sum_k exp({s_k /T_{\text{soft}}})}\}$ \tcp{Take softmax along feature dimension}

\Output{Return the data embedding $\Bar{d}$}

\end{algorithm}

\textbf{(iv) Training Explanation Policies.} In this step, for each cluster $c_j$, using its complementary data set, we train an offline RL agent. We ensure that all the training conditions (algorithm, weight initialization, optimizers, hyperparameters, etc.) are identical to the training of the original RL policy, except for the modification in the training data. We call this newly learned policy as the explanation policy corresponding to cluster $c_j$. We thus get $n_c$ explanation policies at the end of this step. In addition, we compute data embeddings for complementary data sets (Refer to Algorithm~\ref{alg:alternate_policies_data_embeddings}).

\begin{algorithm}[h!]
\footnotesize
\caption{trainExpPolicies}
\label{alg:alternate_policies_data_embeddings}
\SetKwInOut{Input}{Input}
\SetKwInOut{Initialize}{Initialize}
\SetKwInOut{Output}{Output}

\tcc{Train explanation policies \& compute related data embeddings}
\vspace{1mm}
\Input{Offline data$\{\tau_i\}$, Traj. Embeddings $T$, Traj. Clusters $C$, offlineRLAlgo}

\For{$c_j$ in $C$}{
    
    $\{{\tau_i}\}_j \gets \{\tau_i\} - c_j$ \tcp{Compute complementary dataset corresp. to $c_j$}
    
    $T_j \gets$ gatherTrajectoryEmbeddings$(T, \{{\tau_i}\}_j)$ \tcp{Gather corresp. $\tau$ embeds}
    
    Explanation policy, $\pi_j \gets \text{offlineRLAlgo}(\{\tau_i\}_j)$
    
    Complementary data embedding, $\Bar{d}_j \gets$ generateDataEmbedding$(T_j, M, T_{\text{soft}})$
}

\Output{Explanation policies $\{\pi_j\}$, Complementary data embeddings $\{\Bar{d}_j\}$}

\end{algorithm}

\textbf{(v) Cluster Attribution.} In this final step, given a state, we note the actions suggested by all the explanation policies at this state. We then compute the distances of these actions (where we assume a metric over the action space) from the action suggested by the original RL agent at the state. The explanation policies corresponding to the maximum of these distances form the candidate attribution set. For each policy in this candidate attribution set, we compute the distance between its respective complementary data embedding and the data embedding of the entire training data using the Wasserstein metric for capturing distances between softmax simplices~\citep{vallender1974calculation}. We then select the policy that has the smallest data distance and attribute the decision of the RL agent to the cluster corresponding to this policy(Refer to Algorithm~\ref{alg:generating_cluster_attributions}). Our approach comprised of all five steps is summarized in Algorithm~\ref{alg:traj_attribution_offline_rl}.

\begin{algorithm}
\footnotesize
\caption{generateClusterAttribution}
\label{alg:generating_cluster_attributions}
\SetKwInOut{Input}{Input}
\SetKwInOut{Initialize}{Initialize}
\SetKwInOut{Output}{Output}

\tcc{Generating cluster attributions for $a_{\text{orig}} = \pi_{\text{orig}}(s)$}
\vspace{1mm}

\Input{State $s$, Original Policy $\pi_{\text{orig}}$, Explanation Policies $\{\pi_j\}$, Original Data Embedding $\Bar{d}_{\text{orig}}$, Complementary Data Embeddings $\{\Bar{d}_j\}$}

Original action, $a_{\text{orig}} \gets \pi_{\text{orig}}(s)$

Actions suggested by explanation policies, $a_j \gets \pi_j(s)$

$d_{a_{\text{orig}}, a_j} \gets$ calcActionDistance($a_{\text{orig}}, a_j$)\tcp{Compute action distance}

$K \gets \text{argmax}(d_{a_{\text{orig}}, a_j})$\tcp{Get candidate clusters using argmax}

$w_k \gets W_{\text{dist}}(\Bar{d}_{\text{orig}}, \Bar{d}_k)$\tcp{Compute Wasserstein distance b/w complementary data embeddings of candidate clusters \& orig data embedding}

$c_{\text{final}} \gets \text{argmin}(w_k)$\tcp{Choose cluster with min data embedding dist.}

\Output{$c_{\text{final}}$}
\end{algorithm}

\section{Experiments and Results}
\label{sec:res}
Next, we present experimental results to show the effectiveness of our approach in generating trajectory explanations. We address the following key questions: Q1) Do we generate reliable trajectory explanations? (Sec.~\ref{sec:traj_attr_results})  Q2) How does a human understanding of an environment align with trajectories attributed by our algorithm and what is the scope of data attribution techniques? (Sec.~\ref{sec:human_eval})

\subsection{Experimental Setup}
We first describe the environments, models, and metrics designed to study the reliability of our trajectory explanations.

\textbf{RL Environments.} We perform experiments on three environments: i) \textit{Grid-world} (Figure~\ref{fig:gridworld_task_overview}) which has discrete state and action spaces, ii) \textit{Seaquest} from Atari suite which has environments with continuous visual observations and discrete action spaces~\citep{bellemare13arcade}, and iii) \textit{HalfCheetah} from MuJoCo environments which are control environments with continuous state and action spaces~\citep{6386109}.

\textbf{Offline Data and Sequence Encoders.} For grid-world, we collect offline data of 60 trajectories from policy rollout of other RL agents and train an LSTM-based trajectory encoder following the procedure described in trajectory transformer, replacing the transformer with LSTM. For Seaquest, we collect offline data of 717 trajectories from the \href{https://github.com/takuseno/d4rl-atari}{D4RL-Atari repository} and use a \href{https://huggingface.co/edbeeching/decision_transformer_atari}{pre-trained decision transformer} as trajectory encoder. Similarly, for HalfCheetah, we collect offline data of 1000 trajectories from the  \href{https://github.com/Farama-Foundation/D4RL}{D4RL repository}~\citep{fu2020d4rl} and use a \href{https://github.com/jannerm/trajectory-transformer}{pre-trained trajectory transformer} as a trajectory encoder. To cluster high-level skills in long trajectory sequences, we divide the Seaquest trajectories into 30-length sub-trajectories and the HalfCheetah trajectories into 25-length sub-trajectories. These choices were made based on the transformers' input block sizes and the quality of clustering. 

\textbf{Offline RL Training and Data Embedding.} We train the offline RL agents for each environment using the data collected as follows - for grid-world, we use model-based offline RL, and for Seaquest and HalfCheetah, we employ DiscreteSAC~\citep{christodoulou2019soft} and SAC~\citep{haarnoja2018soft}, respectively, using d3rlpy implementations~\citep{seno2021d3rlpy}. We compute data embedding of entire training data for each of the environments. See Appendix~\ref{sec:addl_training_details} for additional training details.

\textbf{Encoding of Trajectories and Clustering.} We encode the trajectory data using sequence encoders and cluster the output trajectory embeddings using the X-means algorithm. More specifically, we obtain 10 trajectory clusters for grid-world, 8 for Seaquest, and 10 for HalfCheetah. These clusters represent meaningful high-level behaviors such as \textit{`falling into the lava'}, \textit{`filling in oxygen'}, \textit{`taking long forward strides'}, etc.  This is discussed in greater detail in Section~\ref{sec:addl_clustering_analysis}.

\textbf{Complementary Data Sets.} We obtain complementary data sets using the aforementioned cluster information and provide 10 complementary data sets for grid-world, 8 for Seaquest, and 10 for HalfCheetah. Next, we compute data embeddings corresponding to these newly formed data sets. 

\textbf{Explanation Policies.} Subsequently, we train explanation policies on the complementary data sets for each environment. The training produces 10 additional policies for grid-world, 8 policies for Seaquest, and 10 policies for HalfCheetah. In summary, we train the original policy on the entire data, obtain data embedding for the entire data, cluster the trajectories and obtain their explanation policies and complementary data embeddings.

\textbf{Trajectory Attribution.} Finally, we attribute a decision made by the original policy for a given state to a trajectory cluster. We choose top-3 trajectories from these attributed clusters by matching the context for the state-action under consideration with trajectories in the cluster in our experiments.

\textbf{Evaluation Metrics.} We compare policies trained on different data using three metrics (deterministic nature of policies is assumed throughout the discussion) -- \textit{1) Initial State Value Estimate}  denoted by $\mathbb{E}(V(s_0))$ which is a measure of expected long-term returns to evaluate offline RL training as described in \citet{paine2020hyperparameter}, \textit{2) Local Mean Absolute Action-Value Difference:} defined as $\mathbb{E}(|\Delta Q_{\pi_{\text{orig}}}|)=\mathbb{E}(|Q_{\pi_{\text{orig}}}(\pi_{\text{orig}}(s)) - Q_{\pi_{\text{orig}}}(\pi_{j}(s))|)$ that measures how original policy perceives suggestions given by explanation policies, and \textit{3) Action Contrast Measure:} a measure of difference in actions suggested by explanation policies and the original action. Here, we use $\mathbb{E}(\mathbbm{1}({\pi_{\text{orig}}(s) \neq \pi_{j}(s)})$ for discrete action space and $\mathbb{E}(({\pi_{\text{orig}}(s) - \pi_{j}(s)})^2)$ for continuous action space. Further, we compute distances between embeddings of original and complementary data sets using Wasserstein metric: $W_{\text{dist}}(\Bar{d}_{\text{orig}}, \Bar{d}_j)$, later normalized to [0, 1]. Finally, the cluster attribution frequency is measured using metric $\mathbb{P}(c_{\text{final}}  = c_{j})$.

\subsection{Trajectory Attribution Results} \label{sec:traj_attr_results}

\begin{figure}[h!]
    \centering
    \includegraphics[width=0.8\textwidth]{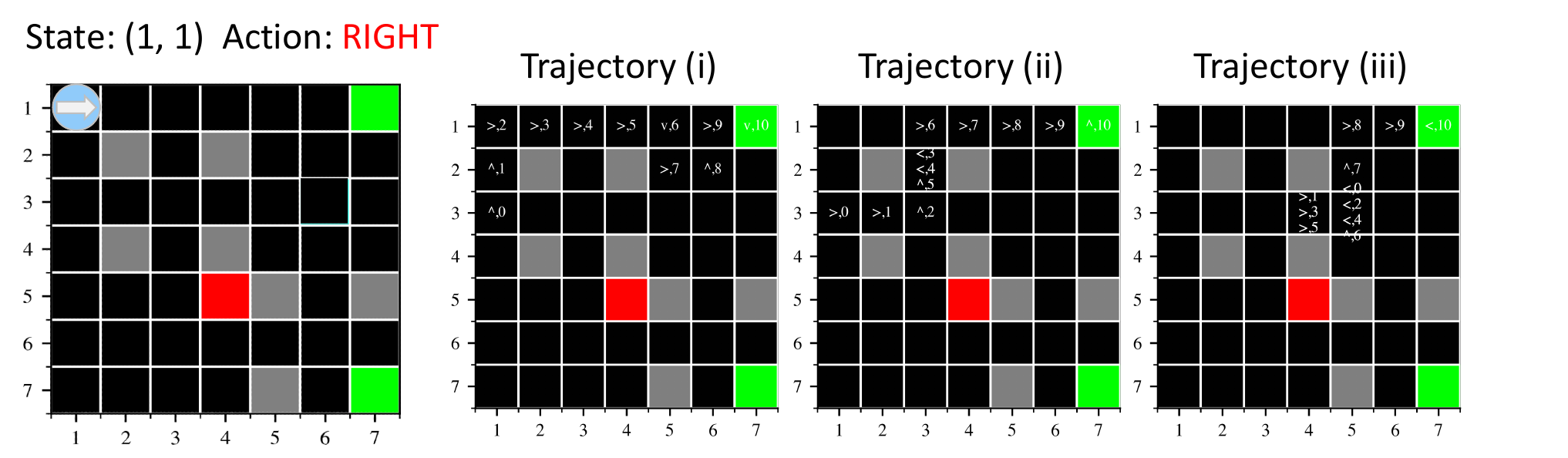}
    \caption{\textbf{Grid-world Trajectory Attribution.} RL agent suggests taking action `right' in grid cell (1,1). This action is attributed to trajectories (i), (ii) and (iii) (We denote gridworld trajectory by annotated $\wedge$,$\vee$,$>$,$<$ arrows for `up', `down', `right', `left' actions respectively, along with the time-step associated with the actions (0-indexed)). We can observe that the RL decisions could be influenced by trajectories distant from the state under consideration, and therefore attributing decisions to trajectories becomes important to understand the decision better.}
    \label{fig:attr_results_gridworld}
\end{figure}

\begin{figure}[h!]
    \centering
    \includegraphics[width=0.8\textwidth]{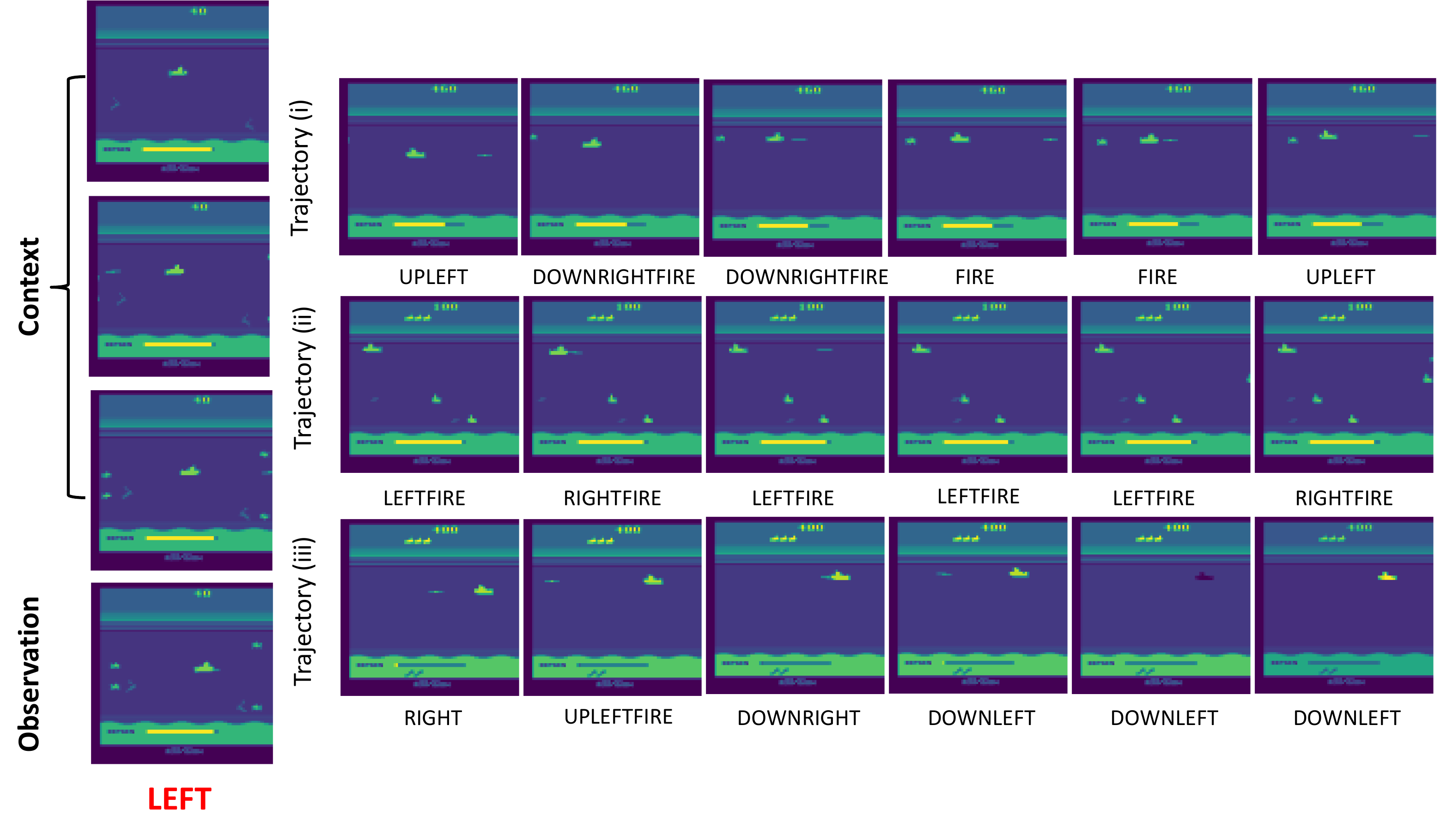}
    \caption{\textbf{Seaquest Trajectory Attribution.} The agent (submarine) decides to take `left' for the given observation under the provided context. Top-3 attributed trajectories are shown on the right (for each training data traj., we show 6 sampled observations and the corresponding actions). As depicted in the attributed trajectories, the action `left' is explained in terms of the agent aligning itself to face the enemies coming from the left end of the frame.}
    \label{fig:attr_results_seaquest}
\end{figure}

\textbf{Qualitative Results.} 
Figure~\ref{fig:attr_results_gridworld} depicts a grid-world state - (1, 1), the corresponding decision by the trained offline RL agent - \textit{`right'}, and attribution trajectories explaining the decision. As we can observe, the decision is influenced not only by trajectory (traj.-i) that goes through (1, 1) but also by other distant trajectories(trajs.-ii, iii). These examples demonstrate that distant experiences (e.g. traj.-iii) could significantly influence the RL agent's decisions, deeming trajectory attribution an essential component of future XRL techniques. Further, Figure~\ref{fig:attr_results_seaquest} shows the Seaquest agent (submarine) suggesting action \textit{`left'} for the given observation in the context of the past few frames. The corresponding attributed trajectories provide insights into how the submarine aligns itself to target enemies coming from the left. Figure~\ref{fig:attr_results_cheetah} shows HalfCheetah observation, the agent suggested action in terms of hinge torques and corresponding attributed trajectories showing runs that influence the suggested set of torques. This is an interesting use-case of trajectory attribution as it explains complicated torques, understood mainly by the domain experts, in terms of the simple semantic intent of \textit{`getting up from the floor'}.

\textbf{Quantitative Analysis.} Tables~\ref{tab:quantitative_results_gridworld},~\ref{tab:quantitative_results_seaquest} and~\ref{tab:quantitative_results_cheetah} present quantitative analysis of the proposed trajectory attribution. The initial state value estimate for the original policy $\pi_{\text{orig}}$ matches or exceeds estimates for explanation policies trained on different complementary data sets in all three environment settings. This indicates that the original policy, having access to all behaviours, is able to outperform other policies that are trained on data lacking information about important behaviours (e.g. grid-world:  reaching a distant goal, Seaquest: fighting in the top-right corner, HalfCheetah: stabilizing the frame while taking strides). Furthermore, local mean absolute action-value difference and action differences turn out to be highly correlated (Tab.~\ref{tab:quantitative_results_seaquest} and~\ref{tab:quantitative_results_cheetah}), i.e., the explanation policies that suggest the most contrasting actions are usually perceived by the original policy as low-return actions. This evidence supports the proposed trajectory algorithm as we want to identify the behaviours which when removed make agent choose actions that are not considered suitable originally. In addition, we provide the distances between the data embeddings in the penultimate column. The cluster attribution distribution is represented in the last column which depicts how RL decisions are dependent on various behaviour clusters. Interestingly, in the case of grid-world, we found that only a few clusters containing information about reaching goals and avoiding lava had the most significant effect on the original RL policy.

\begin{table}[h]
    \centering
    \renewcommand{\arraystretch}{0.9}
    \setlength{\tabcolsep}{1.85pt}
    \caption{\textbf{Quantitative Analysis of Grid-world Trajectory Attribution.} The analysis is provided using 5 metrics. 
    Higher the $\mathbb{E}(V(s_0))$, better is the trained policy. High $\mathbb{E}(|\Delta Q_{\pi_{\text{orig}}}|))$ along with high $\mathbb{E}(\mathbbm{1}({\pi_{\text{orig}}(s) \neq \pi_{j}(s)})$ is desirable. The policies with lower $W_{\text{dist}}(\Bar{d}, \Bar{d}_j)$ and high action contrast are given priority while attribution. The cluster attribution distribution is given in the final column.}
    \begin{tabular}{cccccc}
    \toprule
    \multicolumn{1}{c|}{$\pi$} & \multicolumn{1}{c}{$\mathbb{E}(V(s_0))$} & \multicolumn{1}{c}{$\mathbb{E}(|\Delta Q_{\pi_{\text{orig}}}|))$} & \multicolumn{1}{c}{$\mathbb{E}(\mathbbm{1}({\pi_{\text{orig}}(s) \neq \pi_{j}(s)})$} & \multicolumn{1}{c}{$W_{\text{dist}}(\Bar{d}, \Bar{d}_j)$} & \multicolumn{1}{c}{$\mathbb{P}(c_{\text{final}}  = c_{j})$} \\
    \midrule
    \multicolumn{1}{c|}{orig} & \multicolumn{1}{c}{\textbf{0.3061}} & \multicolumn{1}{c}{-} & \multicolumn{1}{c}{-} & \multicolumn{1}{c}{-} & \multicolumn{1}{c}{-} \\
    \multicolumn{1}{c|}{0} & \multicolumn{1}{c}{0.3055} & \multicolumn{1}{c}{0.0012} & \multicolumn{1}{c}{0.0409} & \multicolumn{1}{c}{\textbf{1.0000}} & \multicolumn{1}{c}{0.0000} \\
    \multicolumn{1}{c|}{1} & \multicolumn{1}{c}{0.3053} & \multicolumn{1}{c}{0.0016} & \multicolumn{1}{c}{0.0409} & \multicolumn{1}{c}{0.0163} & \multicolumn{1}{c}{0.0000} \\
    \multicolumn{1}{c|}{2} & \multicolumn{1}{c}{0.3049} & \multicolumn{1}{c}{0.0289} & \multicolumn{1}{c}{\textbf{0.1429}} & \multicolumn{1}{c}{0.0034} & \multicolumn{1}{c}{0.0000} \\
    \multicolumn{1}{c|}{3} & \multicolumn{1}{c}{0.2857} & \multicolumn{1}{c}{\textbf{0.0710}} & \multicolumn{1}{c}{0.1021} & \multicolumn{1}{c}{0.0111} & \multicolumn{1}{c}{0.3750} \\
    \multicolumn{1}{c|}{4} & \multicolumn{1}{c}{0.2987} & \multicolumn{1}{c}{0.0322} & \multicolumn{1}{c}{\textbf{0.1429}} & \multicolumn{1}{c}{0.0042} & \multicolumn{1}{c}{0.1250} \\
    \multicolumn{1}{c|}{5} & \multicolumn{1}{c}{0.3057} & \multicolumn{1}{c}{0.0393} & \multicolumn{1}{c}{0.0409} & \multicolumn{1}{c}{0.0058} & \multicolumn{1}{c}{0.0000} \\
    \multicolumn{1}{c|}{6} & \multicolumn{1}{c}{0.3046} & \multicolumn{1}{c}{0.0203} & \multicolumn{1}{c}{0.1225} & \multicolumn{1}{c}{0.0005} & \multicolumn{1}{c}{\textbf{0.5000}} \\
    \multicolumn{1}{c|}{7} & \multicolumn{1}{c}{0.3055} & \multicolumn{1}{c}{0.0120} & \multicolumn{1}{c}{0.0205} & \multicolumn{1}{c}{0.0006} & \multicolumn{1}{c}{0.0000} \\
    \multicolumn{1}{c|}{8} & \multicolumn{1}{c}{0.3057} & \multicolumn{1}{c}{0.0008} & \multicolumn{1}{c}{0.0205} & \multicolumn{1}{c}{0.0026} & \multicolumn{1}{c}{0.0000} \\
    \multicolumn{1}{c|}{9} & \multicolumn{1}{c}{0.3046} & \multicolumn{1}{c}{0.0234} & \multicolumn{1}{c}{\textbf{0.1429}} & \multicolumn{1}{c}{0.1745} & \multicolumn{1}{c}{0.0000}\\
    \bottomrule
    \end{tabular}
    \label{tab:quantitative_results_gridworld}
\end{table}

\begin{table}[h]
    \centering
    \renewcommand{\arraystretch}{0.9}
    \setlength{\tabcolsep}{1.85pt}
    \caption{\textbf{Quantitative Analysis of Seaquest Trajectory Attribution.} The analysis is provided using 5 metrics. 
    Higher the $\mathbb{E}(V(s_0))$, better is the trained policy. High $\mathbb{E}(|\Delta Q_{\pi_{\text{orig}}}|))$ along with high $\mathbb{E}(\mathbbm{1}({\pi_{\text{orig}}(s) \neq \pi_{j}(s)})$ is desirable. The policies with lower $W_{\text{dist}}(\Bar{d}, \Bar{d}_j)$ and high action contrast are given priority while attribution. The cluster attribution distribution is given in the final column.}
    \begin{tabular}{cccccc}
    \toprule
    \multicolumn{1}{c|}{$\pi$} & \multicolumn{1}{c}{$\mathbb{E}(V(s_0))$} & \multicolumn{1}{c}{$\mathbb{E}(|\Delta Q_{\pi_{\text{orig}}}|))$} & \multicolumn{1}{c}{$\mathbb{E}(\mathbbm{1}({\pi_{\text{orig}}(s) \neq \pi_{j}(s)})$} & \multicolumn{1}{c}{$W_{\text{dist}}(\Bar{d}, \Bar{d}_j)$} & \multicolumn{1}{c}{$\mathbb{P}(c_{\text{final}}  = c_{j})$} \\
    \midrule
    \multicolumn{1}{c|}{orig} & \multicolumn{1}{c}{85.9977} & \multicolumn{1}{c}{-} & \multicolumn{1}{c}{-} & \multicolumn{1}{c}{-} & \multicolumn{1}{c}{-} \\
    \multicolumn{1}{c|}{0} & \multicolumn{1}{c}{50.9399} & \multicolumn{1}{c}{1.5839} & \multicolumn{1}{c}{0.9249} & \multicolumn{1}{c}{0.4765} & \multicolumn{1}{c}{0.1129} \\
    \multicolumn{1}{c|}{1} & \multicolumn{1}{c}{57.5608} & \multicolumn{1}{c}{1.6352} & \multicolumn{1}{c}{0.8976} & \multicolumn{1}{c}{0.9513} & \multicolumn{1}{c}{0.0484} \\
    \multicolumn{1}{c|}{2} & \multicolumn{1}{c}{66.7369} & \multicolumn{1}{c}{1.5786} & \multicolumn{1}{c}{0.9233} & \multicolumn{1}{c}{\textbf{1.0000}}  & \multicolumn{1}{c}{0.0403} \\
    \multicolumn{1}{c|}{3} & \multicolumn{1}{c}{3.0056} & \multicolumn{1}{c}{\textbf{1.9439}} & \multicolumn{1}{c}{\textbf{0.9395}} & \multicolumn{1}{c}{0.8999} & \multicolumn{1}{c}{0.0323} \\
    \multicolumn{1}{c|}{4} & \multicolumn{1}{c}{58.1854} & \multicolumn{1}{c}{1.5813} & \multicolumn{1}{c}{0.8992} & \multicolumn{1}{c}{0.5532} & \multicolumn{1}{c}{0.0968} \\
    \multicolumn{1}{c|}{5} & \multicolumn{1}{c}{87.3034} & \multicolumn{1}{c}{1.6026} & \multicolumn{1}{c}{0.9254} & \multicolumn{1}{c}{0.2011} & \multicolumn{1}{c}{\textbf{0.3145}} \\
    \multicolumn{1}{c|}{6} & \multicolumn{1}{c}{70.8994} & \multicolumn{1}{c}{1.5501} & \multicolumn{1}{c}{0.9238} & \multicolumn{1}{c}{0.6952} & \multicolumn{1}{c}{0.0968} \\
    \multicolumn{1}{c|}{7} & \multicolumn{1}{c}{\textbf{89.1832}} & \multicolumn{1}{c}{1.5628} & \multicolumn{1}{c}{0.9249} & \multicolumn{1}{c}{0.3090} & \multicolumn{1}{c}{0.2581} \\ \hline
    \end{tabular}
    \label{tab:quantitative_results_seaquest}
\end{table}

We conduct two additional analyses – 1) trajectory attribution on Seaquest trained using Discrete BCQ (Sec.~\ref{sec:seaquest_traj_attribution_bcq}), and 2) Breakout trajectory attribution (Sec.~\ref{sec:breakout_traj_attribution}). In the first one, we find a similar influence of clusters across the decision-making of policies trained under different algorithms. Secondly, in the breakout results, we find clusters with high-level meanings of \textit{`depletion of a life’} and \textit{`right corner shot’} influence decision-making a lot. This is insightful as the two behaviors are highly critical in taking action, one avoids the premature end of the game, and the other is part of the tunneling strategy previously found in ~\cite{greydanus2018visualizing}.

\subsection{Quantifying utility of the Trajectory Attribution: A Human Study} \label{sec:human_eval}
One of the key desiderata of explanations is to provide useful relevant information about the behaviour of complex AI models. To this end, prior works in other domains like vision~\citep{goyal2019counterfactual} and language~\citep{liu-etal-2019-towards-explainable} have conducted human studies to quantify the usefulness of output explanations. Similarly, having established a  straightforward attribution technique, we wish to analyze the utility of the generated attributions and their scope in the real world through a human study. Interestingly, humans possess an explicit understanding of RL gaming environments and can reason about actions at a given state to a satisfactory extent. Leveraging this, we pilot a human study with 10 participants who had a complete understanding of the grid-world navigation environment to quantify the alignment between human knowledge of the environment dynamics with actual factors influencing RL decision-making.

\begin{figure}[H]
    \begin{subfigure}[b]{0.66\textwidth}
         \centering
        \includegraphics[width=0.99\textwidth]{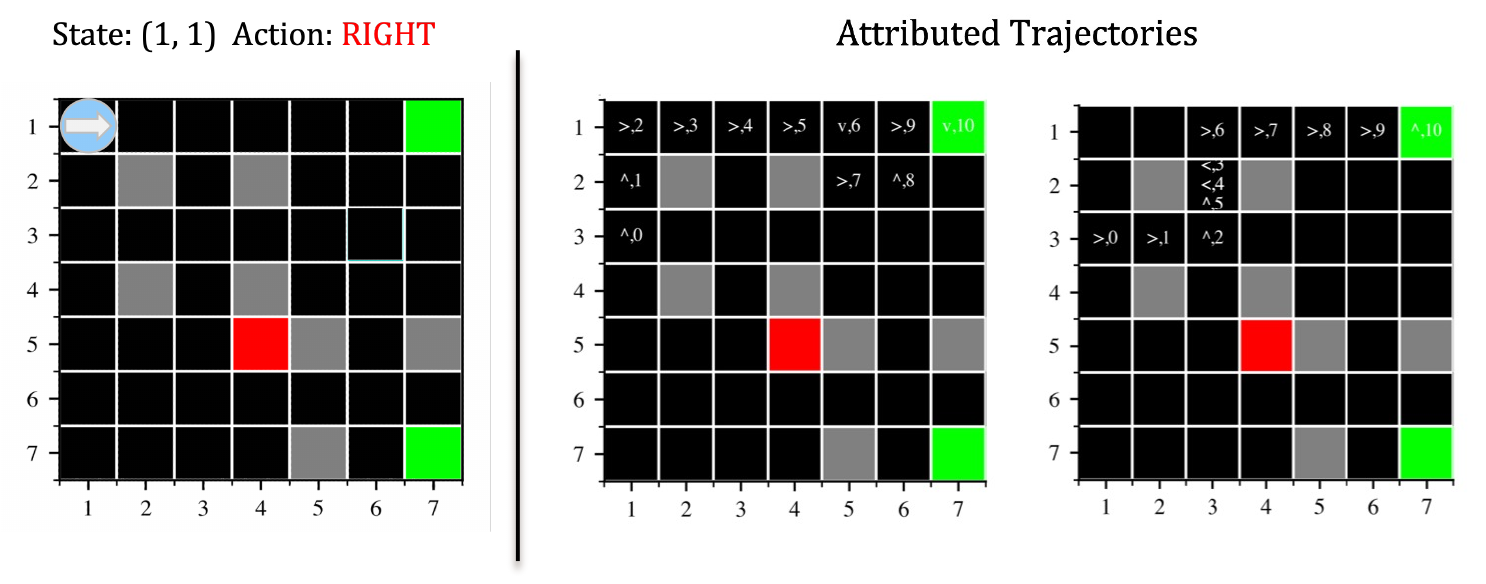}
         \caption{}
         \label{fig:human-study}
    \end{subfigure}
    \begin{subfigure}[b]{0.33\textwidth}
         \centering
         \includegraphics[width=0.99\textwidth]{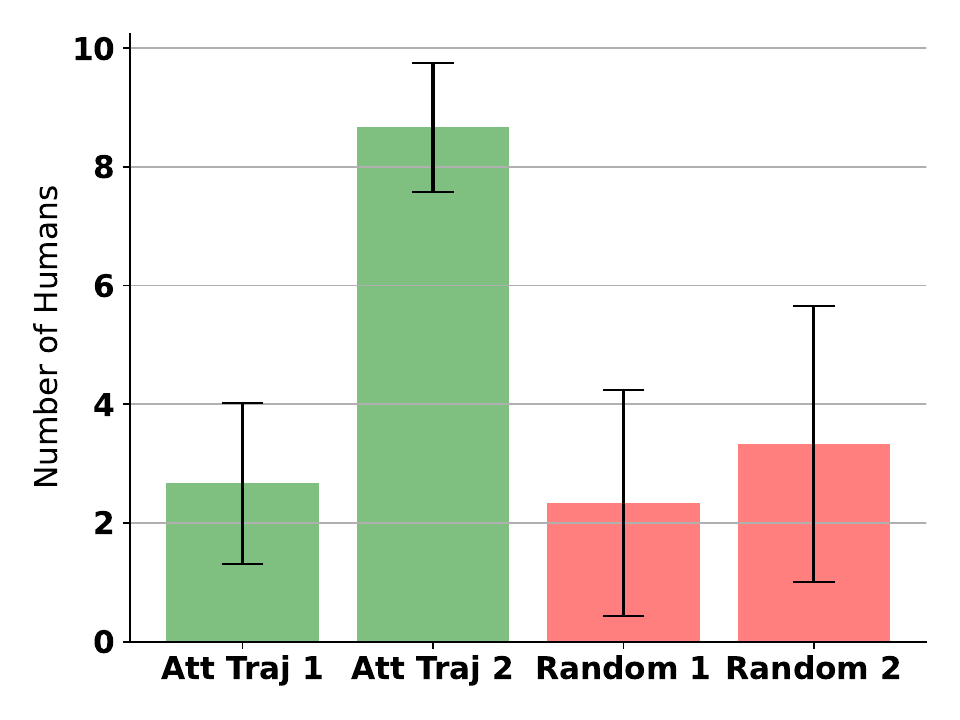}
         \caption{}
         \label{fig:human-stats}
    \end{subfigure}
    \vskip -0.1in
    \caption{\textbf{Column (a):} An example of the human study experiment where users are required to identify the attributed trajectories that best explain the state-action behavior of the agent. \textbf{Column (b):} Results from our human study experiments show a decent alignment of human knowledge of navigation task with actual factors influencing RL decision-making. This underlines the utility as well as the scope of the proposed trajectory attribution explanation method.}
    \label{fig:human-study-full}
\end{figure}

\textbf{Study setup.} For the study, we design two tasks: i) participants need to choose a trajectory that they think \textit{best} explains the action suggested in a grid cell, and ii) participants need to identify \textit{all relevant} trajectories that explain action suggested by RL agent. For instance, in Fig.~\ref{fig:human-study}, we show one instance of the task where the agent is located at (1, 1) and is taking \textit{`right'} and a subset of attributed trajectories for this agent action. In both tasks, in addition to attributions proposed by our technique, we add i) a randomly selected trajectory as an explanation and ii) a trajectory selected from a cluster different from the one attributed by our approach. These additional trajectories aid in identifying human bias toward certain trajectories while understanding the agents’ behavior.

\textbf{Results.} On average, across three studies in Task 1, we found that 70\% of the time, human participants chose trajectories attributed by our proposed method as the best explanation for the agent’s action. On average, across three studies in Task 2, nine participants chose the trajectories generated by our algorithm (Attr Traj 2). In Fig.~\ref{fig:human-stats}, we observe good alignment between human's understanding of  trajectories influencing decision-making involved in grid navigation. Interestingly, the results also demonstrate that not all trajectories generated by our algorithm are considered relevant by humans, and often they are considered as good as a random trajectory (Fig.~\ref{fig:human-stats}; Attr Traj 1). 

In all, as per the Task 1 results, on average 30\% of the time, humans fail to correctly identify the factors influencing an RL decision. Additionally, actual factors driving actions could be neglected by humans while understanding a decision as per Task 2 results. These findings highlight the necessity to have data attribution-based explainability tools to build trust among human stakeholders for handing over the decision-making to RL agents in near future.

\section{Discussion} \label{sec:conc}
In this work, we proposed a novel explanation technique that attributes decisions suggested by an RL agent to trajectories encountered by the agent in the past. We provided an algorithm that enables us to perform trajectory attribution in offline RL. The key idea behind our approach was to encode trajectories using sequence modelling techniques, cluster the trajectories using these embeddings and then study the sensitivity of the original RL agent's policy to the trajectories present in each of these clusters. We demonstrated the utility of our method using experiments in  grid-world, Seaquest and HalfCheetah environments. In the end, we also presented a human evaluation of the results of our method to underline the necessity of trajectory attribution.

The ideas presented in this paper, such as generating trajectory embedding using sequence encoders, and creating an encoding of the set of trajectories, can be extended to other domains of RL. For instance, the trajectory embeddings could be beneficial in recognizing hierarchical behaviour patterns as studied under options theory~\citep{sutton1999between}. Likewise, the data encoding could prove helpful in studying transfer learning~\citep{zhu2020transfer} in RL by utilizing the data embeddings from both the source and target decision-making tasks. From the XRL point of view, we wish to extend our work to online RL settings where the agent constantly collects experiences from the environment.

While our results are highly encouraging, one of the limitations of this work is that there are no established evaluation benchmarks as well as no other works to compare with. We believe that more extensive human studies will help address this.

\section*{Acknowledgements}

We thank anonymous reviewers for their helpful feedback to make this work better. Moreover, NJ acknowledges funding support from NSF IIS-2112471 and NSF CAREER IIS-2141781.

Finally, we wish to dedicate this work to the memory of our dear colleague Georgios Theocharous who is not with us anymore.
While his premature demise has left an unfillable void,  his work has made an indelible mark in the domain of reinforcement learning and in the lives of many researchers. He will forever remain in our memories.




\bibliography{iclr2023_conference}

\begin{thebibliography}{52}
\providecommand{\natexlab}[1]{#1}
\providecommand{\url}[1]{\texttt{#1}}
\expandafter\ifx\csname urlstyle\endcsname\relax
  \providecommand{\doi}[1]{doi: #1}\else
  \providecommand{\doi}{doi: \begingroup \urlstyle{rm}\Url}\fi

\bibitem[{Bellemare} et~al.(2013){Bellemare}, {Naddaf}, {Veness}, and
  {Bowling}]{bellemare13arcade}
M.~G. {Bellemare}, Y.~{Naddaf}, J.~{Veness}, and M.~{Bowling}.
\newblock The arcade learning environment: An evaluation platform for general
  agents.
\newblock \emph{Journal of Artificial Intelligence Research}, 47:\penalty0
  253--279, jun 2013.

\bibitem[Boehnlein et~al.(2021)Boehnlein, Diefenthaler, Fanelli, Hjorth-Jensen,
  Horn, Kuchera, Lee, Nazarewicz, Orginos, Ostroumov, et~al.]{Boehnlein_2022}
Amber Boehnlein, Markus Diefenthaler, Cristiano Fanelli, Morten Hjorth-Jensen,
  Tanja Horn, Michelle~P Kuchera, Dean Lee, Witold Nazarewicz, Kostas Orginos,
  Peter Ostroumov, et~al.
\newblock Artificial intelligence and machine learning in nuclear physics.
\newblock \emph{arXiv preprint arXiv:2112.02309}, 2021.

\bibitem[Briggs(2021)]{briggs_2021}
James Briggs.
\newblock How to get sentence embedding using bert?, Oct 2021.
\newblock URL \url{https://datascience.stackexchange.com/a/103569}.

\bibitem[Chen et~al.(2021)Chen, Lu, Rajeswaran, Lee, Grover, Laskin, Abbeel,
  Srinivas, and Mordatch]{chen2021decisiontransformer}
Lili Chen, Kevin Lu, Aravind Rajeswaran, Kimin Lee, Aditya Grover, Michael
  Laskin, Pieter Abbeel, Aravind Srinivas, and Igor Mordatch.
\newblock Decision transformer: Reinforcement learning via sequence modeling.
\newblock \emph{arXiv preprint arXiv:2106.01345}, 2021.

\bibitem[Choi et~al.(2021)Choi, Kim, Joe, and Gwon]{choi2021evaluation}
Hyunjin Choi, Judong Kim, Seongho Joe, and Youngjune Gwon.
\newblock Evaluation of bert and albert sentence embedding performance on
  downstream nlp tasks.
\newblock In \emph{2020 25th International conference on pattern recognition
  (ICPR)}, pp.\  5482--5487. IEEE, 2021.

\bibitem[Christodoulou(2019)]{christodoulou2019soft}
Petros Christodoulou.
\newblock Soft actor-critic for discrete action settings.
\newblock \emph{arXiv preprint arXiv:1910.07207}, 2019.

\bibitem[Coppens et~al.(2019)Coppens, Efthymiadis, Lenaerts, Now{\'e}, Miller,
  Weber, and Magazzeni]{coppens2019distilling}
Youri Coppens, Kyriakos Efthymiadis, Tom Lenaerts, Ann Now{\'e}, Tim Miller,
  Rosina Weber, and Daniele Magazzeni.
\newblock Distilling deep reinforcement learning policies in soft decision
  trees.
\newblock In \emph{Proceedings of the IJCAI 2019 workshop on explainable
  artificial intelligence}, pp.\  1--6, 2019.

\bibitem[Danesh et~al.(2021)Danesh, Koul, Fern, and Khorram]{danesh2021re}
Mohamad~H Danesh, Anurag Koul, Alan Fern, and Saeed Khorram.
\newblock Re-understanding finite-state representations of recurrent policy
  networks.
\newblock In \emph{International Conference on Machine Learning}, pp.\
  2388--2397. PMLR, 2021.

\bibitem[Fu et~al.(2020)Fu, Kumar, Nachum, Tucker, and Levine]{fu2020d4rl}
Justin Fu, Aviral Kumar, Ofir Nachum, George Tucker, and Sergey Levine.
\newblock D4rl: Datasets for deep data-driven reinforcement learning, 2020.

\bibitem[Fujimoto et~al.(2019{\natexlab{a}})Fujimoto, Conti, Ghavamzadeh, and
  Pineau]{fujimoto2019benchmarking}
Scott Fujimoto, Edoardo Conti, Mohammad Ghavamzadeh, and Joelle Pineau.
\newblock Benchmarking batch deep reinforcement learning algorithms.
\newblock \emph{arXiv preprint arXiv:1910.01708}, 2019{\natexlab{a}}.

\bibitem[Fujimoto et~al.(2019{\natexlab{b}})Fujimoto, Meger, and
  Precup]{fujimoto2019off}
Scott Fujimoto, David Meger, and Doina Precup.
\newblock Off-policy deep reinforcement learning without exploration.
\newblock In \emph{International conference on machine learning}, pp.\
  2052--2062. PMLR, 2019{\natexlab{b}}.

\bibitem[Goyal et~al.(2019)Goyal, Wu, Ernst, Batra, Parikh, and
  Lee]{goyal2019counterfactual}
Yash Goyal, Ziyan Wu, Jan Ernst, Dhruv Batra, Devi Parikh, and Stefan Lee.
\newblock Counterfactual visual explanations.
\newblock In \emph{International Conference on Machine Learning}. PMLR, 2019.

\bibitem[Greydanus et~al.(2018)Greydanus, Koul, Dodge, and
  Fern]{greydanus2018visualizing}
Samuel Greydanus, Anurag Koul, Jonathan Dodge, and Alan Fern.
\newblock Visualizing and understanding atari agents.
\newblock In \emph{International conference on machine learning}, pp.\
  1792--1801. PMLR, 2018.

\bibitem[Haarnoja et~al.(2018)Haarnoja, Zhou, Hartikainen, Tucker, Ha, Tan,
  Kumar, Zhu, Gupta, Abbeel, et~al.]{haarnoja2018soft}
Tuomas Haarnoja, Aurick Zhou, Kristian Hartikainen, George Tucker, Sehoon Ha,
  Jie Tan, Vikash Kumar, Henry Zhu, Abhishek Gupta, Pieter Abbeel, et~al.
\newblock Soft actor-critic algorithms and applications.
\newblock \emph{arXiv preprint arXiv:1812.05905}, 2018.

\bibitem[Iyer et~al.(2018)Iyer, Li, Li, Lewis, Sundar, and
  Sycara]{iyer2018transparency}
Rahul Iyer, Yuezhang Li, Huao Li, Michael Lewis, Ramitha Sundar, and Katia
  Sycara.
\newblock Transparency and explanation in deep reinforcement learning neural
  networks.
\newblock In \emph{Proceedings of the 2018 AAAI/ACM Conference on AI, Ethics,
  and Society}, pp.\  144--150, 2018.

\bibitem[Janner et~al.(2021)Janner, Li, and Levine]{janner2021sequence}
Michael Janner, Qiyang Li, and Sergey Levine.
\newblock Offline reinforcement learning as one big sequence modeling problem.
\newblock In \emph{Advances in Neural Information Processing Systems}, 2021.

\bibitem[Kaelbling et~al.(1998)Kaelbling, Littman, and
  Cassandra]{kaelbling1998planning}
Leslie~Pack Kaelbling, Michael~L Littman, and Anthony~R Cassandra.
\newblock Planning and acting in partially observable stochastic domains.
\newblock \emph{Artificial intelligence}, 101\penalty0 (1-2):\penalty0 99--134,
  1998.

\bibitem[Kidambi et~al.(2020)Kidambi, Rajeswaran, Netrapalli, and
  Joachims]{kidambi2020morel}
Rahul Kidambi, Aravind Rajeswaran, Praneeth Netrapalli, and Thorsten Joachims.
\newblock Morel: Model-based offline reinforcement learning.
\newblock \emph{Advances in neural information processing systems},
  33:\penalty0 21810--21823, 2020.

\bibitem[Korkmaz(2021)]{korkmaz2021investigating}
Ezgi Korkmaz.
\newblock Investigating vulnerabilities of deep neural policies.
\newblock In \emph{Uncertainty in Artificial Intelligence}, pp.\  1661--1670.
  PMLR, 2021.

\bibitem[Kostrikov et~al.(2021)Kostrikov, Fergus, Tompson, and
  Nachum]{kostrikov2021offline}
Ilya Kostrikov, Rob Fergus, Jonathan Tompson, and Ofir Nachum.
\newblock Offline reinforcement learning with fisher divergence critic
  regularization.
\newblock In \emph{International Conference on Machine Learning}, pp.\
  5774--5783. PMLR, 2021.

\bibitem[Koul et~al.(2018)Koul, Greydanus, and Fern]{koul2018learning}
Anurag Koul, Sam Greydanus, and Alan Fern.
\newblock Learning finite state representations of recurrent policy networks.
\newblock \emph{arXiv preprint arXiv:1811.12530}, 2018.

\bibitem[Kumar et~al.(2019)Kumar, Fu, Tucker, and
  Levine]{Kumar2019StabilizingOQ}
Aviral Kumar, Justin Fu, G.~Tucker, and Sergey Levine.
\newblock Stabilizing off-policy q-learning via bootstrapping error reduction.
\newblock In \emph{NeurIPS}, 2019.

\bibitem[Kumar et~al.(2020)Kumar, Zhou, Tucker, and
  Levine]{kumar2020conservative}
Aviral Kumar, Aurick Zhou, George Tucker, and Sergey Levine.
\newblock Conservative q-learning for offline reinforcement learning.
\newblock \emph{Advances in Neural Information Processing Systems},
  33:\penalty0 1179--1191, 2020.

\bibitem[Levine et~al.(2020)Levine, Kumar, Tucker, and
  Fu]{DBLP:journals/corr/abs-2005-01643}
Sergey Levine, Aviral Kumar, George Tucker, and Justin Fu.
\newblock Offline reinforcement learning: Tutorial, review, and perspectives on
  open problems.
\newblock \emph{CoRR}, abs/2005.01643, 2020.
\newblock URL \url{https://arxiv.org/abs/2005.01643}.

\bibitem[Liu et~al.(2019)Liu, Yin, and Wang]{liu-etal-2019-towards-explainable}
Hui Liu, Qingyu Yin, and William~Yang Wang.
\newblock Towards explainable {NLP}: A generative explanation framework for
  text classification.
\newblock In \emph{Proceedings of the 57th Annual Meeting of the Association
  for Computational Linguistics}, pp.\  5570--5581, Florence, Italy, July 2019.
  Association for Computational Linguistics.
\newblock \doi{10.18653/v1/P19-1560}.
\newblock URL \url{https://aclanthology.org/P19-1560}.

\bibitem[Lloyd(1982)]{lloyd1982least}
Stuart Lloyd.
\newblock Least squares quantization in pcm.
\newblock \emph{IEEE transactions on information theory}, 28\penalty0
  (2):\penalty0 129--137, 1982.

\bibitem[Loftus et~al.(2020)Loftus, Filiberto, Li, Balch, Cook, Tighe, Efron,
  Upchurch~Jr, Rashidi, Li, et~al.]{loftus2020decision}
Tyler~J Loftus, Amanda~C Filiberto, Yanjun Li, Jeremy Balch, Allyson~C Cook,
  Patrick~J Tighe, Philip~A Efron, Gilbert~R Upchurch~Jr, Parisa Rashidi,
  Xiaolin Li, et~al.
\newblock Decision analysis and reinforcement learning in surgical
  decision-making.
\newblock \emph{Surgery}, 168\penalty0 (2):\penalty0 253--266, 2020.

\bibitem[Madumal et~al.(2020)Madumal, Miller, Sonenberg, and
  Vetere]{madumal2020explainable}
Prashan Madumal, Tim Miller, Liz Sonenberg, and Frank Vetere.
\newblock Explainable reinforcement learning through a causal lens.
\newblock In \emph{Proceedings of the AAAI conference on artificial
  intelligence}, volume~34, pp.\  2493--2500, 2020.

\bibitem[Mnih et~al.(2013)Mnih, Kavukcuoglu, Silver, Graves, Antonoglou,
  Wierstra, and Riedmiller]{mnih2013playing}
Volodymyr Mnih, Koray Kavukcuoglu, David Silver, Alex Graves, Ioannis
  Antonoglou, Daan Wierstra, and Martin Riedmiller.
\newblock Playing atari with deep reinforcement learning.
\newblock \emph{arXiv preprint arXiv:1312.5602}, 2013.

\bibitem[Nguyen et~al.(2021)Nguyen, Kim, and Nguyen]{nguyen2021effectiveness}
Giang Nguyen, Daeyoung Kim, and Anh Nguyen.
\newblock The effectiveness of feature attribution methods and its correlation
  with automatic evaluation scores.
\newblock \emph{Advances in Neural Information Processing Systems},
  34:\penalty0 26422--26436, 2021.

\bibitem[Novikov(2019)]{Novikov2019}
Andrei Novikov.
\newblock {PyClustering}: Data mining library.
\newblock \emph{Journal of Open Source Software}, 4\penalty0 (36):\penalty0
  1230, apr 2019.
\newblock \doi{10.21105/joss.01230}.
\newblock URL \url{https://doi.org/10.21105/joss.01230}.

\bibitem[Paine et~al.(2020)Paine, Paduraru, Michi, Gulcehre, Zolna, Novikov,
  Wang, and de~Freitas]{paine2020hyperparameter}
Tom~Le Paine, Cosmin Paduraru, Andrea Michi, Caglar Gulcehre, Konrad Zolna,
  Alexander Novikov, Ziyu Wang, and Nando de~Freitas.
\newblock Hyperparameter selection for offline reinforcement learning.
\newblock \emph{arXiv preprint arXiv:2007.09055}, 2020.

\bibitem[Park et~al.(2018)Park, Kim, Kang, Chung, and Choi]{park2018sequence}
Seong~Hyeon Park, ByeongDo Kim, Chang~Mook Kang, Chung~Choo Chung, and Jun~Won
  Choi.
\newblock Sequence-to-sequence prediction of vehicle trajectory via lstm
  encoder-decoder architecture.
\newblock In \emph{2018 IEEE Intelligent Vehicles Symposium (IV)}, pp.\
  1672--1678. IEEE, 2018.

\bibitem[Pawlowski et~al.(2020)Pawlowski, Coelho~de Castro, and
  Glocker]{pawlowski2020deep}
Nick Pawlowski, Daniel Coelho~de Castro, and Ben Glocker.
\newblock Deep structural causal models for tractable counterfactual inference.
\newblock \emph{Advances in Neural Information Processing Systems},
  33:\penalty0 857--869, 2020.

\bibitem[Pelleg et~al.(2000)Pelleg, Moore, et~al.]{pelleg2000x}
Dan Pelleg, Andrew~W Moore, et~al.
\newblock X-means: Extending k-means with efficient estimation of the number of
  clusters.
\newblock In \emph{Icml}, volume~1, pp.\  727--734, 2000.

\bibitem[Puiutta \& Veith(2020)Puiutta and Veith]{puiutta2020explainable}
Erika Puiutta and Eric Veith.
\newblock Explainable reinforcement learning: A survey.
\newblock In \emph{International cross-domain conference for machine learning
  and knowledge extraction}, pp.\  77--95. Springer, 2020.

\bibitem[Puri et~al.(2019)Puri, Verma, Gupta, Kayastha, Deshmukh,
  Krishnamurthy, and Singh]{puri2019explain}
Nikaash Puri, Sukriti Verma, Piyush Gupta, Dhruv Kayastha, Shripad Deshmukh,
  Balaji Krishnamurthy, and Sameer Singh.
\newblock Explain your move: Understanding agent actions using specific and
  relevant feature attribution.
\newblock \emph{arXiv preprint arXiv:1912.12191}, 2019.

\bibitem[Puterman(2014)]{puterman2014markov}
Martin~L Puterman.
\newblock \emph{Markov decision processes: discrete stochastic dynamic
  programming}.
\newblock John Wiley \& Sons, 2014.

\bibitem[Reed et~al.(2022)Reed, Zolna, Parisotto, Colmenarejo, Novikov,
  Barth-Maron, Gimenez, Sulsky, Kay, Springenberg, et~al.]{reed2022generalist}
Scott Reed, Konrad Zolna, Emilio Parisotto, Sergio~Gomez Colmenarejo, Alexander
  Novikov, Gabriel Barth-Maron, Mai Gimenez, Yury Sulsky, Jackie Kay,
  Jost~Tobias Springenberg, et~al.
\newblock A generalist agent.
\newblock \emph{arXiv preprint arXiv:2205.06175}, 2022.

\bibitem[Schulman et~al.(2017)Schulman, Wolski, Dhariwal, Radford, and
  Klimov]{schulman2017proximal}
John Schulman, Filip Wolski, Prafulla Dhariwal, Alec Radford, and Oleg Klimov.
\newblock Proximal policy optimization algorithms.
\newblock \emph{arXiv preprint arXiv:1707.06347}, 2017.

\bibitem[Silver et~al.(2017)Silver, Schrittwieser, Simonyan, Antonoglou, Huang,
  Guez, Hubert, Baker, Lai, Bolton, et~al.]{silver2017mastering}
David Silver, Julian Schrittwieser, Karen Simonyan, Ioannis Antonoglou, Aja
  Huang, Arthur Guez, Thomas Hubert, Lucas Baker, Matthew Lai, Adrian Bolton,
  et~al.
\newblock Mastering the game of go without human knowledge.
\newblock \emph{nature}, 550\penalty0 (7676):\penalty0 354--359, 2017.

\bibitem[Sutton \& Barto(2018)Sutton and Barto]{sutton2018reinforcement}
Richard~S Sutton and Andrew~G Barto.
\newblock \emph{Reinforcement learning: An introduction}.
\newblock MIT press, 2018.

\bibitem[Sutton et~al.(1999)Sutton, Precup, and Singh]{sutton1999between}
Richard~S Sutton, Doina Precup, and Satinder Singh.
\newblock Between mdps and semi-mdps: A framework for temporal abstraction in
  reinforcement learning.
\newblock \emph{Artificial intelligence}, 112\penalty0 (1-2):\penalty0
  181--211, 1999.

\bibitem[Takuma~Seno(2021)]{seno2021d3rlpy}
Michita~Imai Takuma~Seno.
\newblock d3rlpy: An offline deep reinforcement library.
\newblock In \emph{NeurIPS 2021 Offline Reinforcement Learning Workshop},
  December 2021.

\bibitem[Todorov et~al.(2012)Todorov, Erez, and Tassa]{6386109}
Emanuel Todorov, Tom Erez, and Yuval Tassa.
\newblock Mujoco: A physics engine for model-based control.
\newblock In \emph{2012 IEEE/RSJ International Conference on Intelligent Robots
  and Systems}, pp.\  5026--5033, 2012.
\newblock \doi{10.1109/IROS.2012.6386109}.

\bibitem[Vallender(1974)]{vallender1974calculation}
SS~Vallender.
\newblock Calculation of the wasserstein distance between probability
  distributions on the line.
\newblock \emph{Theory of Probability \& Its Applications}, 18\penalty0
  (4):\penalty0 784--786, 1974.

\bibitem[Vaswani et~al.(2017)Vaswani, Shazeer, Parmar, Uszkoreit, Jones, Gomez,
  Kaiser, and Polosukhin]{vaswani2017attention}
Ashish Vaswani, Noam Shazeer, Niki Parmar, Jakob Uszkoreit, Llion Jones,
  Aidan~N Gomez, {\L}ukasz Kaiser, and Illia Polosukhin.
\newblock Attention is all you need.
\newblock \emph{Advances in neural information processing systems}, 30, 2017.

\bibitem[Verma et~al.(2018)Verma, Murali, Singh, Kohli, and
  Chaudhuri]{verma2018programmatically}
Abhinav Verma, Vijayaraghavan Murali, Rishabh Singh, Pushmeet Kohli, and Swarat
  Chaudhuri.
\newblock Programmatically interpretable reinforcement learning.
\newblock In \emph{International Conference on Machine Learning}, pp.\
  5045--5054. PMLR, 2018.

\bibitem[Yu et~al.(2020)Yu, Thomas, Yu, Ermon, Zou, Levine, Finn, and
  Ma]{yu2020mopo}
Tianhe Yu, Garrett Thomas, Lantao Yu, Stefano Ermon, James~Y Zou, Sergey
  Levine, Chelsea Finn, and Tengyu Ma.
\newblock Mopo: Model-based offline policy optimization.
\newblock \emph{Advances in Neural Information Processing Systems},
  33:\penalty0 14129--14142, 2020.

\bibitem[Zaheer et~al.(2017)Zaheer, Kottur, Ravanbakhsh, P{\'{o}}czos,
  Salakhutdinov, and Smola]{DBLP:journals/corr/ZaheerKRPSS17}
Manzil Zaheer, Satwik Kottur, Siamak Ravanbakhsh, Barnab{\'{a}}s P{\'{o}}czos,
  Ruslan Salakhutdinov, and Alexander~J. Smola.
\newblock Deep sets.
\newblock \emph{CoRR}, abs/1703.06114, 2017.
\newblock URL \url{http://arxiv.org/abs/1703.06114}.

\bibitem[Zhu et~al.(2018)Zhu, Guo, Xu, Liao, Yang, and Huang]{zhu2018group}
Feiyun Zhu, Jun Guo, Zheng Xu, Peng Liao, Liu Yang, and Junzhou Huang.
\newblock Group-driven reinforcement learning for personalized mhealth
  intervention.
\newblock In \emph{International Conference on Medical Image Computing and
  Computer-Assisted Intervention}, pp.\  590--598. Springer, 2018.

\bibitem[Zhu et~al.(2020)Zhu, Lin, and Zhou]{zhu2020transfer}
Zhuangdi Zhu, Kaixiang Lin, and Jiayu Zhou.
\newblock Transfer learning in deep reinforcement learning: A survey.
\newblock \emph{arXiv preprint arXiv:2009.07888}, 2020.

\end{thebibliography}
\bibliographystyle{iclr2023_conference}

\clearpage
\appendix
\section{Appendix} \label{sec:appendix}
\subsection{Overview of Proposed Trajectory Attribution}
The following is an overview of our proposed 5-step trajectory attribution algorithm.

\begin{algorithm}[H]
\caption{Trajectory Attribution in Offline RL}
\label{alg:traj_attribution_offline_rl}
\SetKwInOut{Input}{Input}
\SetKwInOut{Initialize}{Initialize}
\SetKwInOut{Output}{Output}

\Input{Offline Data \{$\tau_i$\}, States needing explanation $\mathcal{S}_{\text{exp}}$, Sequence Encoder $E$, offlineRLAlgo, clusteringAlgo, Normalizing constant $M$, Softmax Temperature $T_{\text{soft}}$}

\tcc{Train original offline RL policy}
$\pi_{\text{orig}} \gets \text{offlineRLAlgo}(\{\tau_i\})$

\tcc{Encode individual trajectories}
$T$ = encodeTrajectories(\{$\tau_i$\}, $E$) \tcp{Algo.~\ref{alg:trajectory_encoding}}

\tcc{Cluster the trajectories}
$C \gets$ clusterTrajectories(T, clusteringAlgo) \tcp{Algo.~\ref{alg:cluster_trajs}}

\tcc{Compute data embedding for the entire dataset}
$\Bar{d}_{\text{orig}}$ = generateDataEmbedding(T, $M, T_{\text{soft}})$ \tcp{Algo.~\ref{alg:data_encoding}}

\tcc{Generate explanation policies and their corresponding complementary data embeddings}
$\{\pi_j\}$, $\{\Bar{d}_j\}$ $\gets$ trainExpPolicies($\{\tau_i\}$, T, $C$, offlineRLAlgo)\tcp{Algo.~\ref{alg:alternate_policies_data_embeddings}}

\tcc{Attributing policy decisions for given set of states}
\For{$s \in \mathcal{S}_{\text{exp}}$}{
    
    $c_{\text{final}} \gets $ generateClusterAttribution($s, \pi_{\text{orig}}, \{\pi_j\}, \Bar{d}_{\text{orig}}, {\Bar{d}_j}$) \tcp{Algo.~\ref{alg:generating_cluster_attributions}}
    
    *Optionally, select top N trajectories in the cluster $c_{\text{final}}$ using a pre-defined criteria.
}
 
\end{algorithm}

\subsection{Grid-world Environment Details}
\begin{figure}[H]
    \centering
    \includegraphics[width=0.5\textwidth]{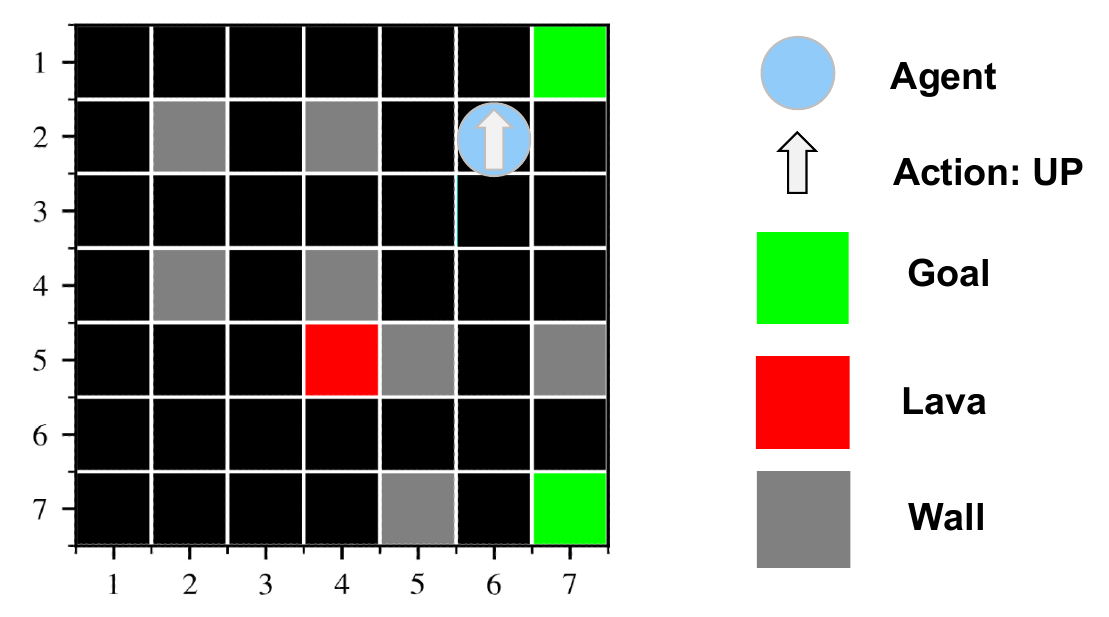}
    \caption{\textbf{Overview of the Grid-world Environment.}
The aim of the agent is to reach any of the goal states (green squares) by avoiding lava (red square) and going around the impenetrable walls (grey squares). The reward for reaching the goal is +1; if the agent falls into the lava, it is -1. For any other transitions, the agent receives -0.1. The agent is allowed to take up, down, left or right as the action.}
    \label{fig:gridworld_task_overview}
\end{figure}

\subsection{Addtional Training Details}\label{sec:addl_training_details}
\begin{enumerate}
    \item \textbf{Seaquest Atari Environment} -- We employed Discrete SAC to train the original policy along with explanation policies, where the training was performed until saturation in the performance. We used the critic learning rate of $3\times10^{-4}$ and the actor learning rate of $3\times10^{-4}$ with a batch size of 256. The trainings were performed parallelly on a single Nvidia-A100 GPU hardware.
    \item \textbf{HalfCheetah MuJoCo Environment} -- We used SAC to train the original policy as well as explanation policies where we trained the agents until training performance saturated. We again used the critic learning rate of $3\times10^{-4}$ and the actor learning rate of $3\times10^{-4}$ with a batch size of 512. The policy trainings were performed parallelly on a single Nvidia-A100 GPU.
    
\end{enumerate}



\subsection{Clustering Analysis} \label{sec:addl_clustering_analysis}

\begin{figure}[h]
     \centering
     \begin{subfigure}[b]{0.32\textwidth}
         \centering
         \includegraphics[width=\textwidth]{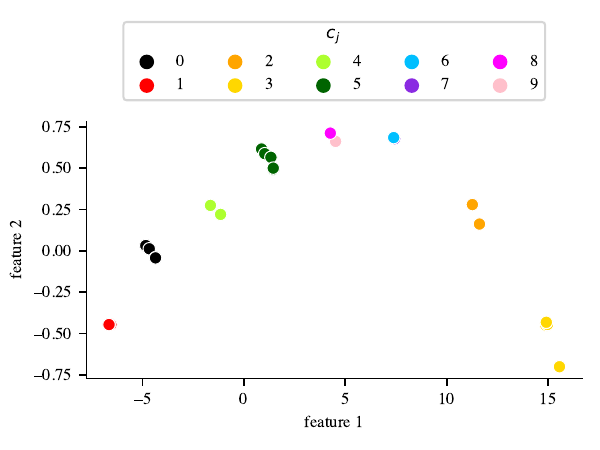}
         \caption{Grid-world}
         \label{fig:traj_clustering_grid}
     \end{subfigure}
     \hfill
     \begin{subfigure}[b]{0.32\textwidth}
         \centering
         \includegraphics[width=\textwidth]{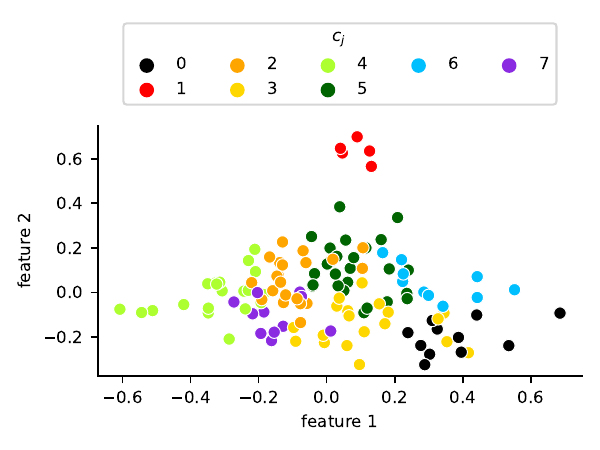}
         \caption{Seaquest}
         \label{fig:traj_clustering_seaquest}
     \end{subfigure}
     \hfill
     \begin{subfigure}[b]{0.32\textwidth}
         \centering
         \includegraphics[width=\textwidth]{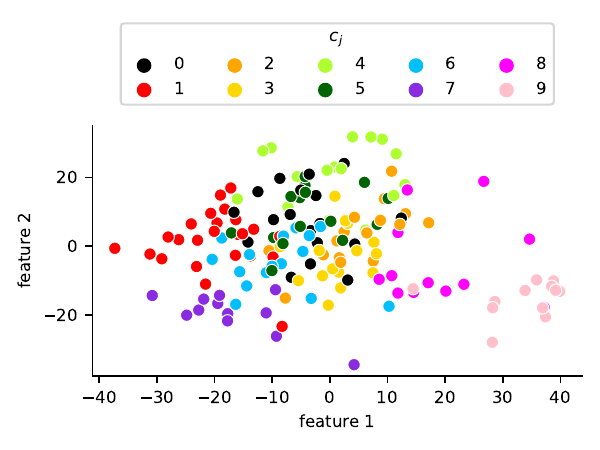}
         \caption{HalfCheetah}
         \label{fig:traj_clustering_cheetah}
     \end{subfigure}
        \caption{\textbf{PCA Plot depicting Clusters of Trajectory Embeddings} for a) Grid-world, b) Seaquest, and c) HalfCheetah. We find that these clusters represent semantically meaningful high-level behaviours.}
        \label{fig:traj_clustering}
\end{figure}

\begin{figure}[H]
    \centering
    \includegraphics[width=\textwidth]{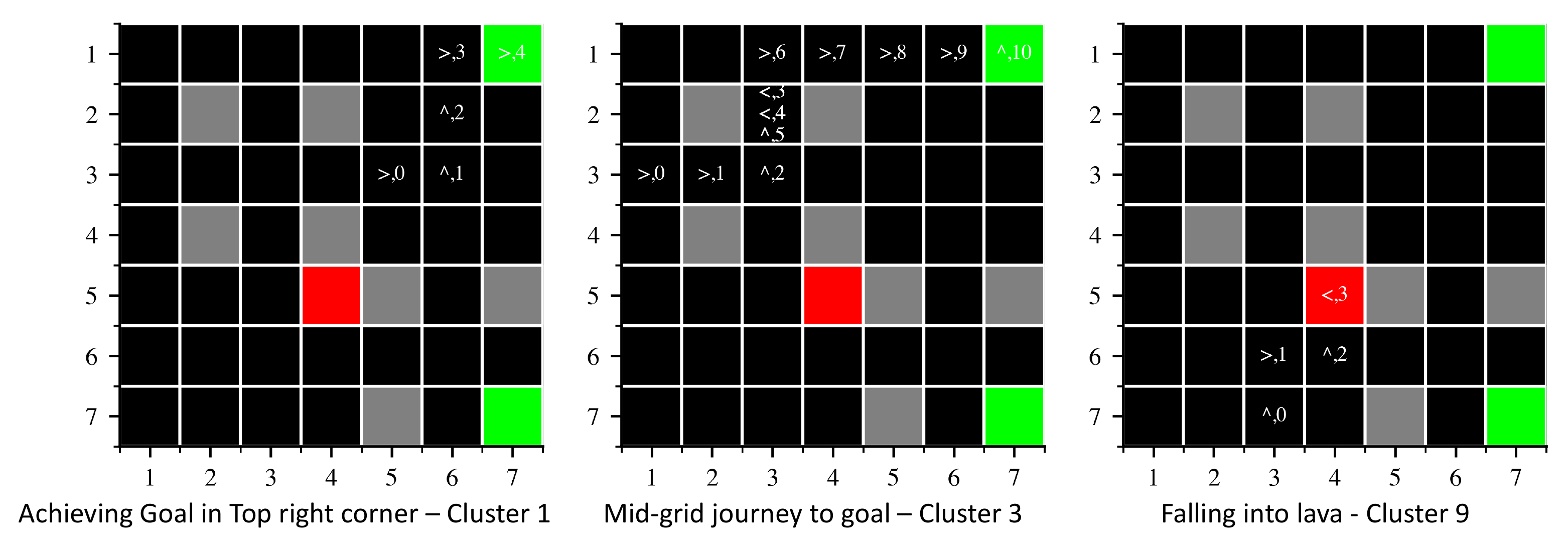}
    \caption{\textbf{Cluster Behaviours for Grid-world.} The figure shows 3 example high-level behaviours along with the action description and id of the cluster representing such behaviour.}
    \label{fig:cluster_behaviours_grid}
\end{figure}

\begin{figure}[H]
    \centering
    \includegraphics[width=0.8\textwidth]{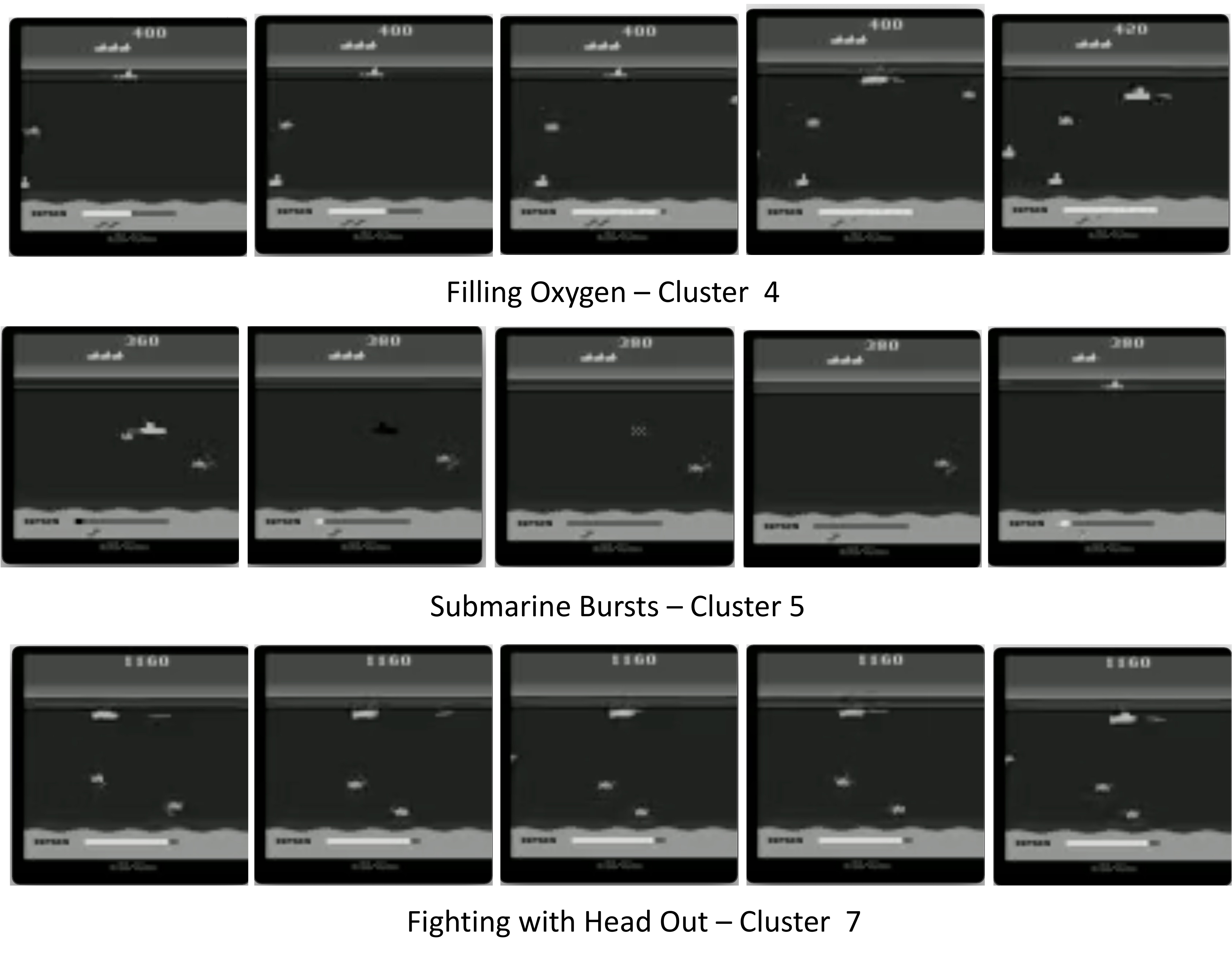}
    \caption{\textbf{High-level Behaviours found in clusters for Seaquest} formed using trajectory embeddings produced using decision transformer. The figure shows 3 example high-level behaviours along with the action description and id of the cluster representing such behaviour.}
    \label{fig:cluster_behaviours_seaquest}
\end{figure}

\begin{figure}[H]
    \centering
    \includegraphics[width=0.9\textwidth]{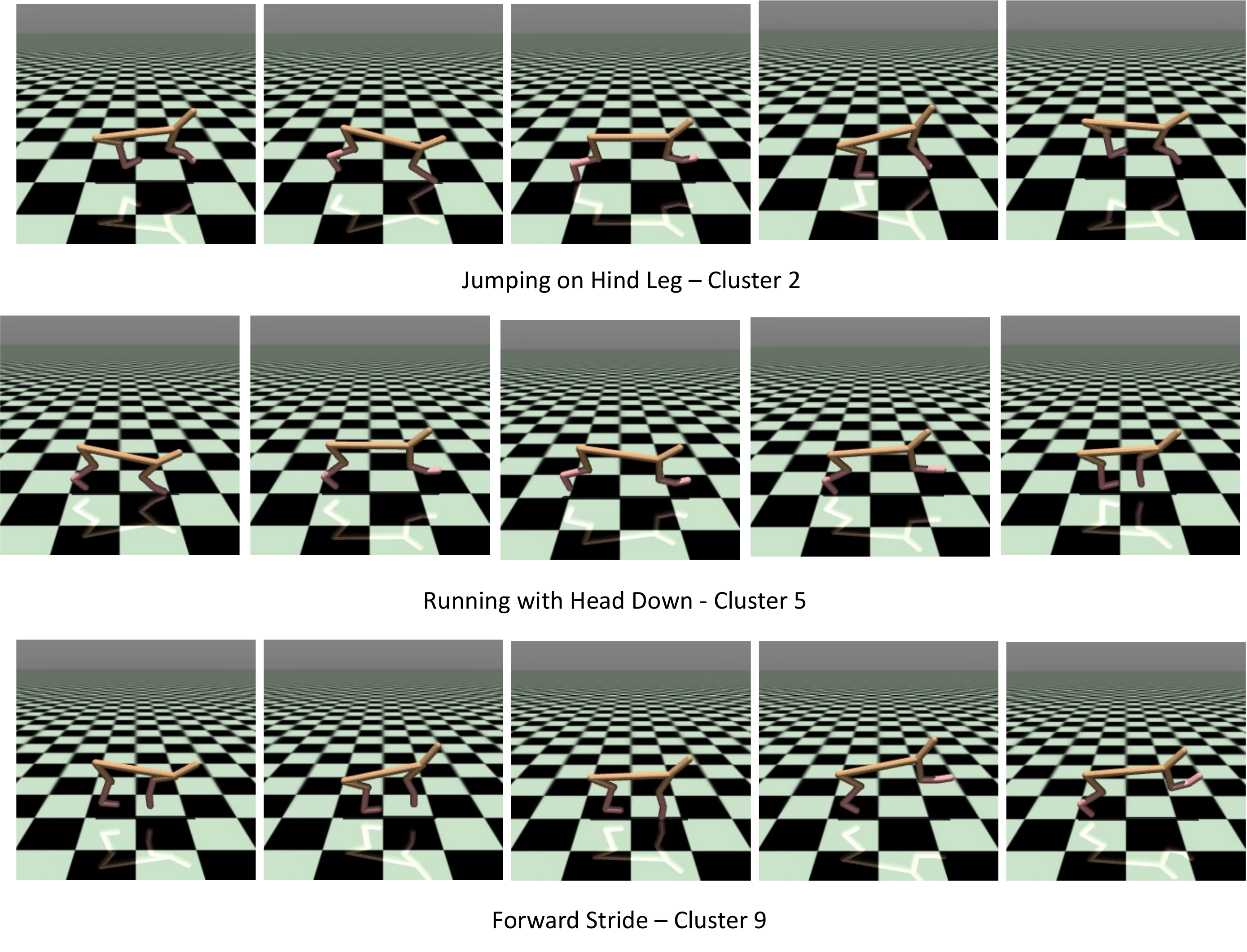}
    \caption{\textbf{High-level Behaviours found in clusters  for HalfCheetah} formed using trajectory embeddings produced using trajectory transformer. The figure shows 3 example high-level behaviours along with the action description and id of the cluster representing such behaviour.}
    \label{fig:cluster_behaviours_cheetah}
\end{figure}

We observe that the trajectory embeddings obtained from the sequence encoders when clustered together demonstrate characteristic high-level behaviours. For instance, in the case of grid-world (Refer~\ref{fig:cluster_behaviours_grid}), the clusters comprise semantically similar trajectories where the agent demonstrates behaviours such as `\textit{falling into the lava}', `\textit{achieving the goal in the first quadrant}', `\textit{mid-grid journey to the goal}', etc. For Seaquest~\ref{fig:cluster_behaviours_seaquest}, we obtain trajectory clusters that represent high-level behaviours such as `\textit{filling in oxygen}', `\textit{fighting along the surface}', `\textit{submarine bursting due to collision with enemy}', etc.  and for HalfCheetah in Fig.~\ref{fig:cluster_behaviours_cheetah}, we obtain trajectory clusters that represent high-level actions such as `\textit{taking long forward strides}', `\textit{jumping on hind leg}', `\textit{running with head down}', etc. 

Although these results look quite promising, in this work we mainly focus on trajectory attribution that leverages these findings. In future, we wish to analyse the trajectory embeddings and the behaviour patterns in greater detail.

\subsection{HalfCheetah Trajectory Attribution Results}
Due to space constraints, we present the qualitative and quantitative results for the HalfCheetah environment here.

\begin{figure}[H]
    \centering
    \includegraphics[width=\textwidth]{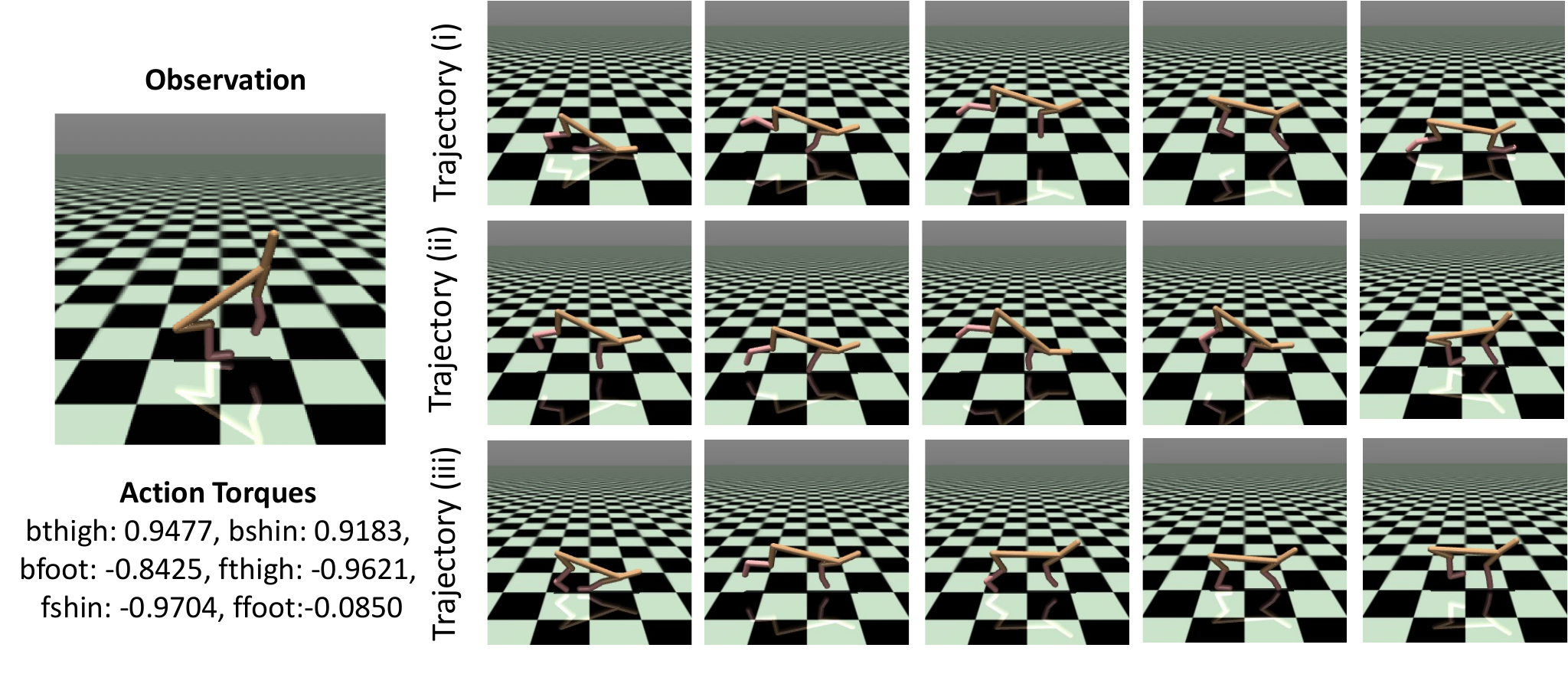}
    \caption{\textbf{HalfCheetah Trajectory Attribution.} The figure shows agent suggesting torques on different hinges for current position of the cheetah frame. The decision is influenced by the runs of cheetah shown on the right (we show 5 sampled frames for one trajectory). The attributed trajectories explain the torques in terms of the cheetah getting up from the floor. (Context is not shown here because unlike Seaquest environment, HalfCheetah decisions are made directly on a given observation~\citep{puterman2014markov, kaelbling1998planning}.)}
    \label{fig:attr_results_cheetah}
\end{figure}

\begin{table}[h]
    \centering
    \renewcommand{\arraystretch}{0.9}
    \setlength{\tabcolsep}{1.85pt}
    \caption{\textbf{Quantitative Analysis of HalfCheetah Trajectory Attribution.} The analysis is provided using 5 metrics. Higher the $\mathbb{E}(V(s_0))$, better is the trained policy. High $\mathbb{E}(|\Delta Q_{\pi_{\text{orig}}}|))$ along with high $\mathbb{E}(({\pi_{\text{orig}}(s) - \pi_{j}(s)})^2)$ is desirable. The policies with lower $W_{\text{dist}}(\Bar{d}, \Bar{d}_j)$ and high action contrast are given priority while attribution. The cluster attribution distribution is given in the final column.}
    \begin{tabular}{cccccc}
    \toprule
    \multicolumn{1}{c|}{} & \multicolumn{5}{c}{Performance Metrics}\\
    \multicolumn{1}{c|}{$\pi$} & \multicolumn{1}{c}{$\mathbb{E}(V(s_0))$} & \multicolumn{1}{c}{$\mathbb{E}(|\Delta Q_{\pi_{\text{orig}}}|))$} & \multicolumn{1}{c}{$\mathbb{E}(({\pi_{\text{orig}}(s) - \pi_{j}(s)})^2)$} & \multicolumn{1}{c}{$W_{\text{dist}}(\Bar{d}, \Bar{d}_j)$} & \multicolumn{1}{c}{$\mathbb{P}(c_{\text{final}}  = c_{j})$} \\
    \midrule
    \multicolumn{1}{c|}{orig} & \multicolumn{1}{c}{131.5449} & \multicolumn{1}{c}{-} & \multicolumn{1}{c}{-} & \multicolumn{1}{c}{-} & \multicolumn{1}{c}{-} \\
    \multicolumn{1}{c|}{0} & \multicolumn{1}{c}{127.1652} & \multicolumn{1}{c}{0.5667} & \multicolumn{1}{c}{0.6359} & \multicolumn{1}{c}{0.2822} & \multicolumn{1}{c}{0.0143} \\
    \multicolumn{1}{c|}{1} & \multicolumn{1}{c}{118.5663} & \multicolumn{1}{c}{0.4796} & \multicolumn{1}{c}{0.5633} & \multicolumn{1}{c}{0.0396} & \multicolumn{1}{c}{0.0214} \\
    \multicolumn{1}{c|}{2} & \multicolumn{1}{c}{122.0661} & \multicolumn{1}{c}{0.6904} & \multicolumn{1}{c}{0.9366} & \multicolumn{1}{c}{0.0396} & \multicolumn{1}{c}{0.1464} \\
    \multicolumn{1}{c|}{3} & \multicolumn{1}{c}{133.4590} & \multicolumn{1}{c}{0.5360} & \multicolumn{1}{c}{0.6611} & \multicolumn{1}{c}{0.0396} & \multicolumn{1}{c}{0.1250}\\
    \multicolumn{1}{c|}{4} & \multicolumn{1}{c}{118.3447} & \multicolumn{1}{c}{0.5622} & \multicolumn{1}{c}{0.6194} & \multicolumn{1}{c}{\textbf{1.0000}} & \multicolumn{1}{c}{0.0964}\\
    \multicolumn{1}{c|}{5} & \multicolumn{1}{c}{\textbf{138.7517}} & \multicolumn{1}{c}{0.6439} & \multicolumn{1}{c}{0.8262} & \multicolumn{1}{c}{0.0316} & \multicolumn{1}{c}{0.0893}\\
    \multicolumn{1}{c|}{6} & \multicolumn{1}{c}{120.7088} & \multicolumn{1}{c}{0.4740} & \multicolumn{1}{c}{0.4803} & \multicolumn{1}{c}{0.8813} & \multicolumn{1}{c}{0.0214}\\
    \multicolumn{1}{c|}{7} & \multicolumn{1}{c}{135.6848} & \multicolumn{1}{c}{0.5154} & \multicolumn{1}{c}{0.5489} & \multicolumn{1}{c}{0.0394} & \multicolumn{1}{c}{\textbf{0.2036}}\\
    \multicolumn{1}{c|}{8} & \multicolumn{1}{c}{113.3490} & \multicolumn{1}{c}{0.7826} & \multicolumn{1}{c}{1.0528} & \multicolumn{1}{c}{0.7067} & \multicolumn{1}{c}{0.1214}\\
    \multicolumn{1}{c|}{9} & \multicolumn{1}{c}{83.6211} & \multicolumn{1}{c}{\textbf{0.9702}} & \multicolumn{1}{c}{\textbf{1.3453}} & \multicolumn{1}{c}{0.0264} & \multicolumn{1}{c}{0.1607}\\
    \bottomrule
    \end{tabular}
    \label{tab:quantitative_results_cheetah}
\end{table}

\subsection{Additional Attribution Results} \label{sec:addl_attr_results}

\begin{figure}[H]
    \centering
    \includegraphics[width=0.9\textwidth]{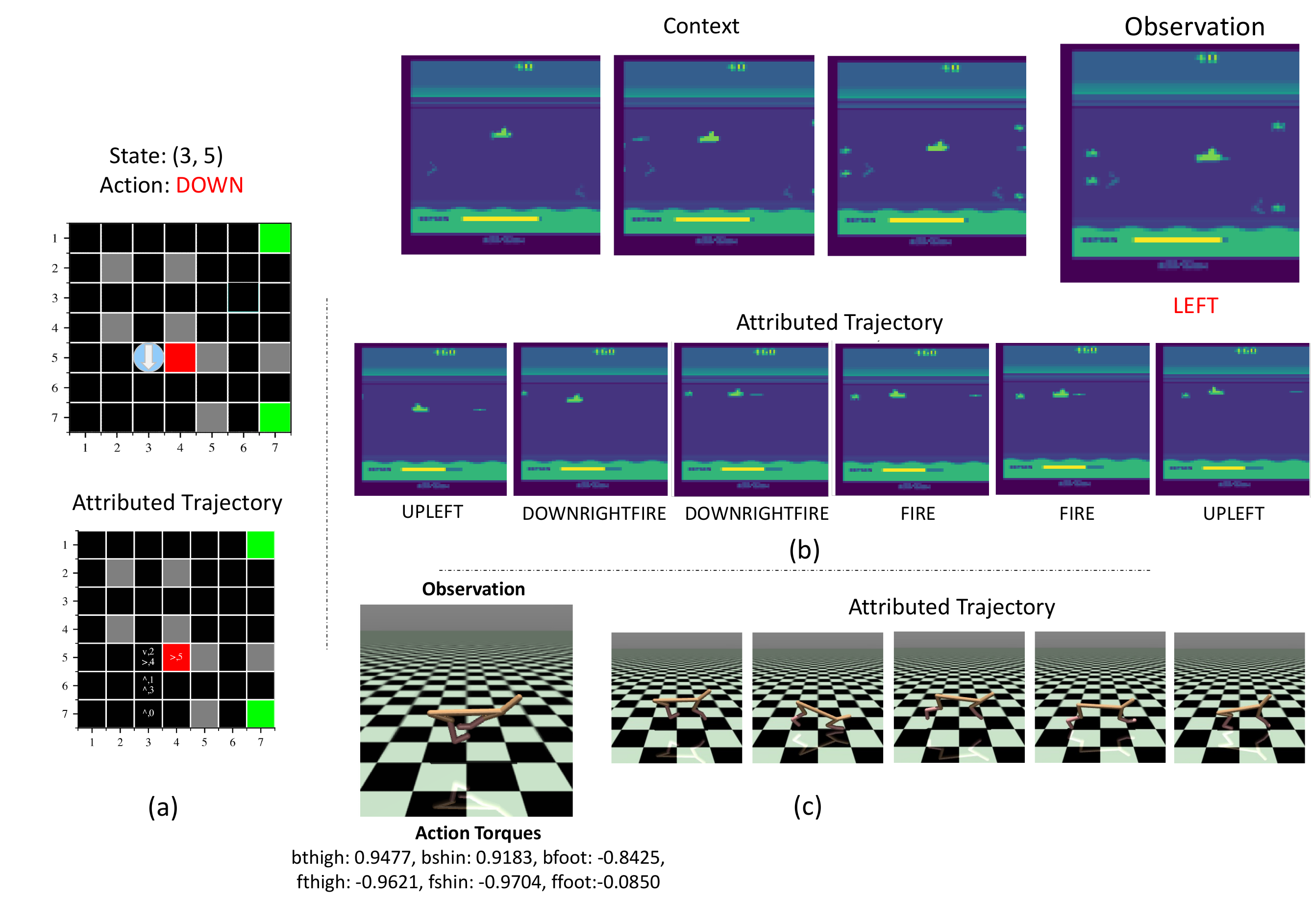}
    \caption{\textbf{Additional Trajectory Attribution Results.} Here we show randomly chosen  trajectory from the top-3 attributed trajectories.a) Grid-world agent suggests taking `DOWN' in cell (3,5) due to the attributed trajectory leading to lava. b) Seaquest agent suggests taking  `LEFT' and the corresponding attributed trajectory. c) HalfCheetah agent suggests a particular set of torques and the run found responsible for the same is shown side-by-side.}
    \label{fig:addl_attr_results}
\end{figure}


\subsection{Trajectory Attribution across algorithms}
\label{sec:seaquest_traj_attribution_bcq}

\begin{figure}[H]
    \centering
    \includegraphics[width=0.75\textwidth]{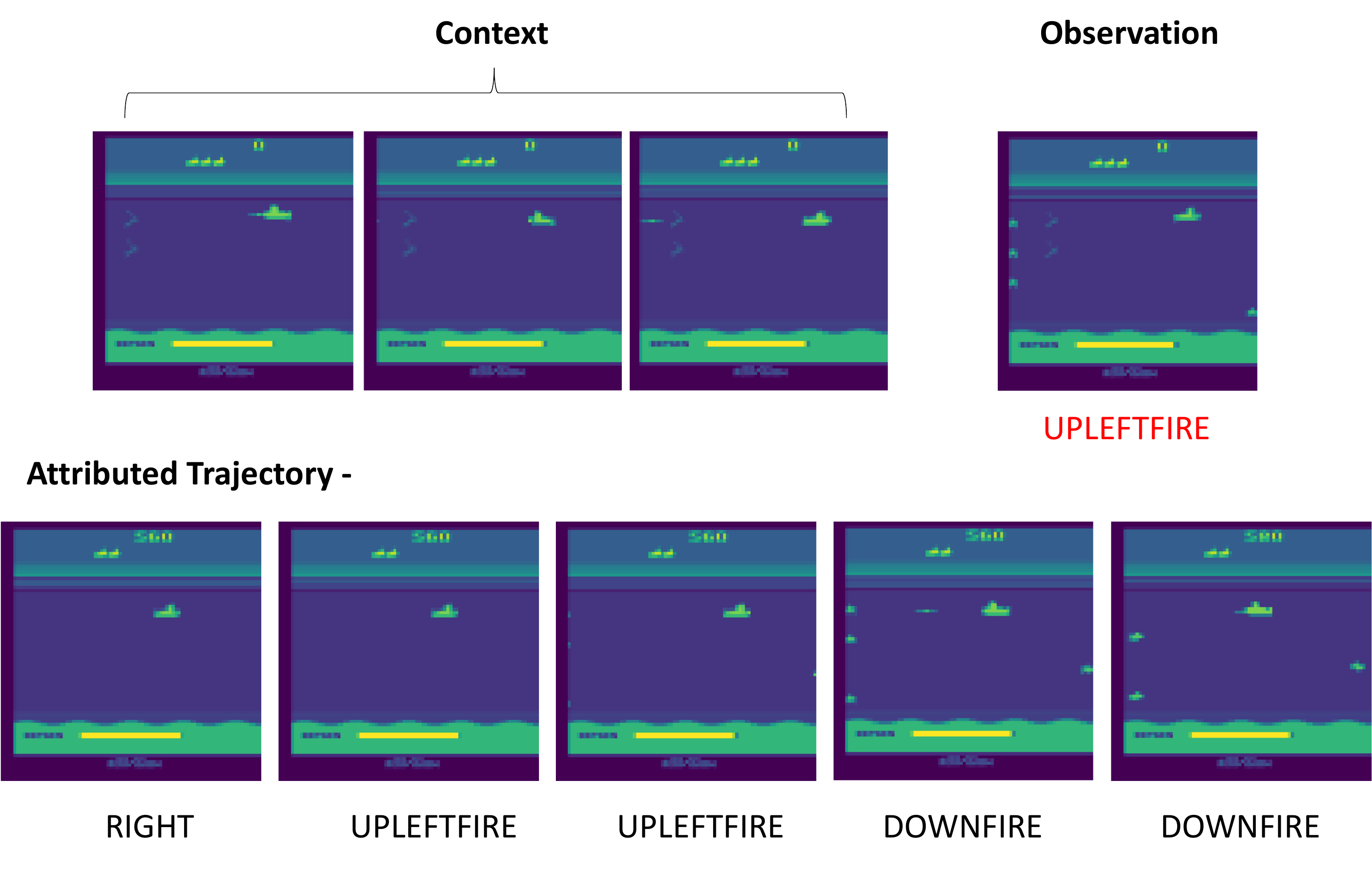}
    \caption{\textbf{Trajectory Attribution for DiscreteBCQ-trained Seaquest Agent.} The action 'UPLEFTFIRE' is explained by our algorithm in terms of corresponding attributed trajectory. The action helps align agent to face enemies from left (akin to Fig.~\ref{fig:attr_results_seaquest}).}
    \label{fig:traj_attr_results_bcq}
\end{figure}

In Sec.~\ref{sec:traj_attr_results}, we perform trajectory attribution on Seaquest environment trained using Discrete SAC algorithm. Here, we show results of our attribution algorithm in identifying influential trajectories for agents trained on same data but with different RL algorithm. Specifically, we choose Discrete Batch Constrained Q-Learning~\citep{fujimoto2019off,fujimoto2019benchmarking}  to train a Seaquest policy. 

Fig.~\ref{fig:traj_attr_results_bcq} depicts a qualitative explanation generated using our algorithm for DiscreteBCQ trained agent. Table~\ref{tab:quantitative_results_seaquest_bcq} gives quantitative numbers associated with attributions performed in this setting. It is quite interesting to note that our proposed algorithm assigns similar importance to various clusters as done in Table~\ref{tab:quantitative_results_seaquest}. That is, we find that certain behaviours in the data agnostic to the algorithm used for training play similar role in determining final execution policy. Thus, we find that our algorithm is generalizable  and reliable enough to provide consistent insights across various RL algorithms.

\begin{table}[H]
    \centering
    \renewcommand{\arraystretch}{0.9}
    \setlength{\tabcolsep}{1.85pt}
    \caption{\textbf{Analysis of Trajectory Attribution for DiscreteBCQ-trained Seaquest Agent.}}
    \begin{tabular}{cccccc}
    \toprule
    \multicolumn{1}{c|}{$\pi$} & \multicolumn{1}{c}{$\mathbb{E}(V(s_0))$} & \multicolumn{1}{c}{$\mathbb{E}(|\Delta Q_{\pi_{\text{orig}}}|))$} & \multicolumn{1}{c}{$\mathbb{E}(\mathbbm{1}({\pi_{\text{orig}}(s) \neq \pi_{j}(s)})$} & \multicolumn{1}{c}{$W_{\text{dist}}(\Bar{d}, \Bar{d}_j)$} & \multicolumn{1}{c}{$\mathbb{P}(c_{\text{final}}  = c_{j})$} \\
    \midrule
    \multicolumn{1}{c|}{orig} & \multicolumn{1}{c}{\textbf{1.3875}} & \multicolumn{1}{c}{-} & \multicolumn{1}{c}{-} & \multicolumn{1}{c}{-} & \multicolumn{1}{c}{-} \\
    \multicolumn{1}{c|}{0} & \multicolumn{1}{c}{0.9619} & \multicolumn{1}{c}{0.1309} & \multicolumn{1}{c}{0.9249} & \multicolumn{1}{c}{0.4765} & \multicolumn{1}{c}{0.1025} \\
    \multicolumn{1}{c|}{1} & \multicolumn{1}{c}{0.5965} & \multicolumn{1}{c}{\textbf{0.1380}} & \multicolumn{1}{c}{0.8976} & \multicolumn{1}{c}{0.9513} & \multicolumn{1}{c}{0.0256} \\
    \multicolumn{1}{c|}{2} & \multicolumn{1}{c}{1.0157} & \multicolumn{1}{c}{0.1325} & \multicolumn{1}{c}{0.9233} & \multicolumn{1}{c}{\textbf{1.0000}}  & \multicolumn{1}{c}{0.00854} \\
    \multicolumn{1}{c|}{3} & \multicolumn{1}{c}{1.1270} & \multicolumn{1}{c}{0.1323} & \multicolumn{1}{c}{\textbf{0.9395}} & \multicolumn{1}{c}{0.8999} & \multicolumn{1}{c}{0.0769} \\
    \multicolumn{1}{c|}{4} & \multicolumn{1}{c}{1.2243} & \multicolumn{1}{c}{0.1280} & \multicolumn{1}{c}{0.8992} & \multicolumn{1}{c}{0.5532} & \multicolumn{1}{c}{0.1025} \\
    \multicolumn{1}{c|}{5} & \multicolumn{1}{c}{1.2143} & \multicolumn{1}{c}{0.1367} & \multicolumn{1}{c}{0.9254} & \multicolumn{1}{c}{0.2011} & \multicolumn{1}{c}{\textbf{0.3248}} \\
    \multicolumn{1}{c|}{6} & \multicolumn{1}{c}{0.9752} & \multicolumn{1}{c}{0.1334} & \multicolumn{1}{c}{0.9238} & \multicolumn{1}{c}{0.6952} & \multicolumn{1}{c}{0.1196} \\
    \multicolumn{1}{c|}{7} & \multicolumn{1}{c}{1.1229} & \multicolumn{1}{c}{0.1352} & \multicolumn{1}{c}{0.9249} & \multicolumn{1}{c}{0.3090} & \multicolumn{1}{c}{0.2393} \\ \hline
    \end{tabular}
    \label{tab:quantitative_results_seaquest_bcq}
\end{table}

\subsection{Trajectory Attribution on Atari Breakout Environment}
\label{sec:breakout_traj_attribution}
\begin{figure}[H]
    \centering
    \includegraphics[width=0.75\textwidth]{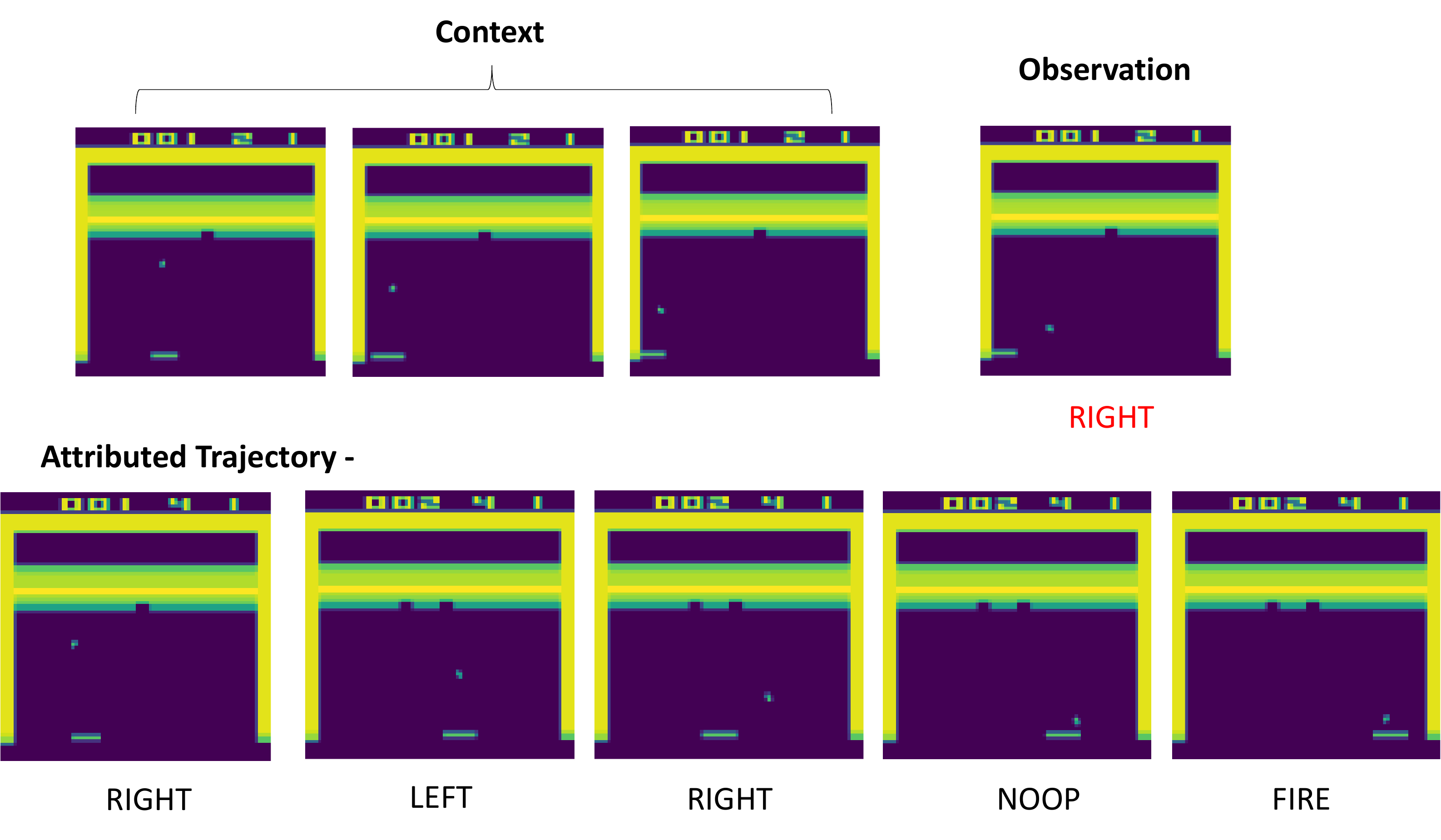}
    \caption{\textbf{Trajectory Attribution for DiscreteBCQ-trained Breakout Agent.} The agent proposes taking `RIGHT' in the given observation frame. The corresponding attribution result shows how the ball coming from left would be played if moved to right.}
    \label{fig:traj_attr_results_breakout_bcq}
\end{figure}

We present attribution results on additional environment of Atari Breakout~\cite{bellemare13arcade} trained using Discrete BCQ. Fig.~\ref{fig:traj_attr_results_breakout_bcq} shows an instance of qualitative result and table~\ref{tab:quantitative_results_breakout_bcq} gives the number for overall attributions performed on Breakout.

\begin{figure}[H]
    \centering
    \includegraphics[width=0.9\textwidth]{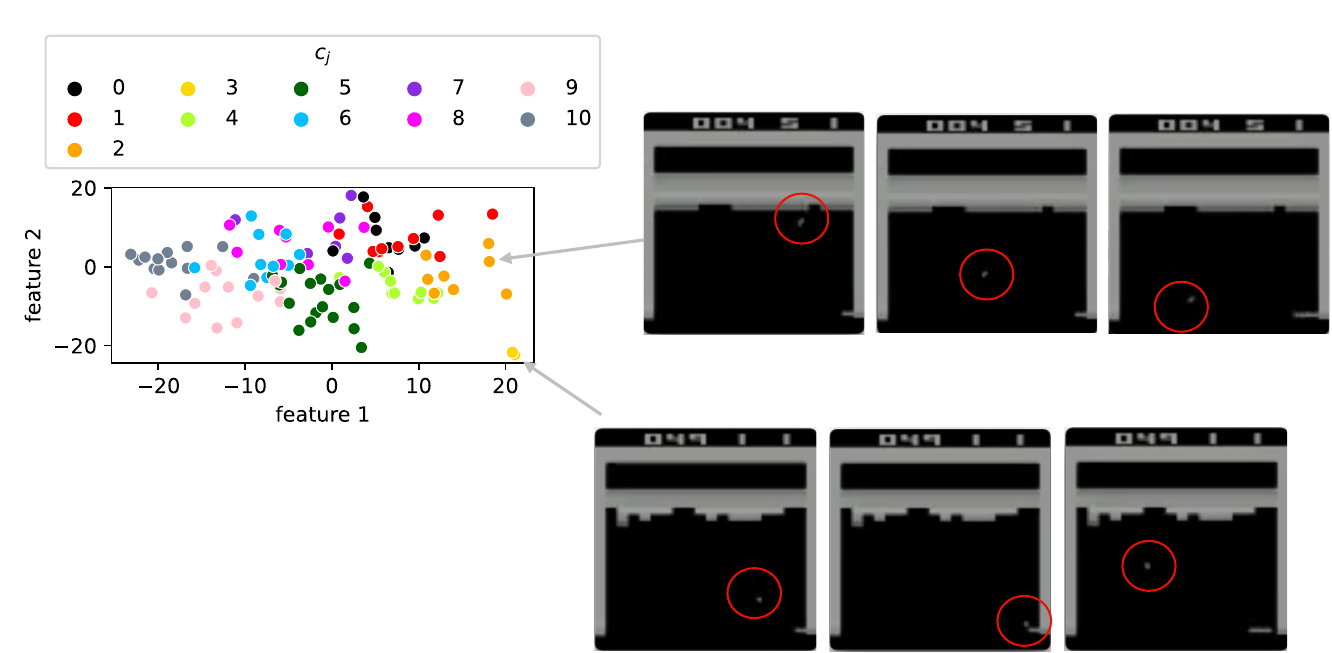}
    \caption{\textbf{Breakout Trajectory Clusters.} The figure shows a PCA plot of breakout trajectories clustered into 11 clusters. Our method identifies cluster 2 (`corner shot from the right') and cluster 3 (`depletion of life') as the most important high-level behaviours in the data for learning.}

    \label{fig:traj_clusters_breakout}
\end{figure}

\begin{table}[h]
    \centering
    \renewcommand{\arraystretch}{0.9}
    \setlength{\tabcolsep}{1.85pt}
    \caption{\textbf{Quantitative Analysis of Trajectory Attribution for DiscreteBCQ-trained Breakout Agent.} We identify that clusters 2 and 3 representing `corner shots from right' and `depletion of a life',  impact the decision-making significantly~\ref{fig:traj_clusters_breakout}. This is insightful given how important these behaviours are in general, the first one shows how to avoid ending the game prematurely, and the second one is a well-known strategy in Breakout for playing at the end of the right frame to break the walls on the top left for creating a tunnel.}
    \begin{tabular}{c|ccccc}
\toprule
$\pi$ & $\mathbb{E}(V(s_0))$ & $\mathbb{E}(|\Delta Q_{\pi_{\text{orig}}}|))$ & $\mathbb{E}(\mathbbm{1}({\pi_{\text{orig}}(s) \neq \pi_{j}(s)})$ & $W_{\text{dist}}(\Bar{d}, \Bar{d}_j)$ & $\mathbb{P}(c_{\text{final}} = c_{j})$  \\ 
\midrule
orig  & 1.4570               & -                                             & -                                                                & -                                     & -                                       \\
0     & 1.1877               & 0.0972                                        & 0.7469                                                           & 0.8828                                & 0.0000                                  \\
1     & \textbf{1.5057}      & 0.0990                                        & 0.7317                                                           & 0.2046                                & 0.1428                                  \\
2     & 1.2107               & 0.0983                                        & 0.7405                                                           & 0.1676                                & 0.2619                                  \\
3     & 1.3946               & 0.0930                                        & 0.6687                                                           & 0.1417                                & \textbf{0.3095}                         \\
4     & 1.4533               & 0.1043                                        & 0.7225                                                           & 0.3827                                & 0.0476                                  \\
5     & 1.4678               & 0.1030                                        & 0.7310                                                           & 0.5339                                & 0.0000                                  \\
6     & 1.1719               & 0.1022                                        & 0.7322                                                           & \textbf{1.0000}                       & 0.0000                                  \\
7     & 1.3493               & \textbf{0.1092}                               & 0.7225                                                           & 0.3935                                & 0.0000                                  \\
8     & 1.2775               & 0.0916                                        & \textbf{0.7604}                                                  & 0.6999                                & 0.04761                                 \\
9     & 1.3773               & 0.0956                                        & 0.7496                                                           & 0.5700                                & 0.04761                                 \\
10    & 1.4351               & 0.0998                                        & 0.7520                                                           & 0.3005                                & 0.1428                                  \\
\hline
\end{tabular}
    \label{tab:quantitative_results_breakout_bcq}
\end{table}

\end{document}